\documentclass[10pt,twocolumn,letterpaper]{article}

\usepackage{wacv}

\usepackage{multirow}

\graphicspath{{}}

\newcommand{\name}{VideoRouteAudit\xspace}

\definecolor{wacvblue}{rgb}{0.21,0.49,0.74}
\usepackage[breaklinks,colorlinks,allcolors=wacvblue]{hyperref}
\hypersetup{
  pdftitle={Where Do Reported Long-Video QA Gains Come From? A Controlled Audit of Frame Selection, Evidence Injection, and Expert Routing},
  pdfauthor={Shih-Yao Lin}
}

\def\wacvPaperID{2752}
\def\confName{WACV}
\def\confYear{2027}

\title{Where Do Reported Long-Video QA Gains Come From?\\
       A Controlled Audit of Frame Selection, Evidence Injection,
       and Expert Routing}

\author{Shih-Yao Lin\\
Independent Researcher\\
\texttt{mikeslin.research@gmail.com}
}

\begin{document}
\maketitle

\begin{abstract}
Long-video question answering is usually studied as frame selection:
which few frames should a vision--language model see? We introduce
\name{}, a controlled audit that separates this choice from two
alternatives: injecting non-visual evidence and routing to another
answering model. It holds the backbone, frame budget, decoding, prompt and
grader fixed while varying one decision at a time. On the full
Video-MME long bucket ($n{=}900$, $K{=}8$), no tested selector
significantly improves over uniform sampling: the largest tested
selection-only estimate is $+1.9$\,pt ($95\%$ CI $[-1.2,+5.0]$) and our
cascade is $+1.2$\,pt $[-1.7,+4.1]$; all are equivalent to uniform
within $\pm5$\,pt. Yet three of four matched contrasts are up to
$3.8\times$ larger on commonly used $n{=}200$ subsets, one reverses
sign, and subsampling the full bucket reverses the full-bucket effect's
sign in $35.8\%$ of draws. The audit also corrects our own first
results: the anomaly-domain ``$+$ASR'' arms injected no text, so the
pooled $+4.7$\,pt gain was selection plus a development prefix; with
transcripts, ASR lowers XD-Violence accuracy by $10.0$\,pt when appended
after the question and not at all when placed before it, and UCF-Crime
by $13.0$\,pt either way, while on Video-MME placing it first adds about
$4$ ($7$ on Qwen2-VL) pt to uniform sampling and to the cascade alike
(post hoc, unadjusted). Swapping the answering model produces the
largest contrast ($+15.1$\,pt on MUSIC-AVQA, $+33.3$\,pt on its worst
type), but the same audio-visual expert with its soundtrack silenced
keeps $92\%$ of the gain ($69\%$ with a mismatched one): mostly a model swap. Code, per-item predictions and scripts that
recompute every statistic are available at
\url{https://github.com/mikeslin-ir/long-video-qa-audit}.
\end{abstract}

\section{Introduction}
\label{sec:intro}

\begin{figure}[!t]
\centering
\includegraphics[width=\columnwidth]{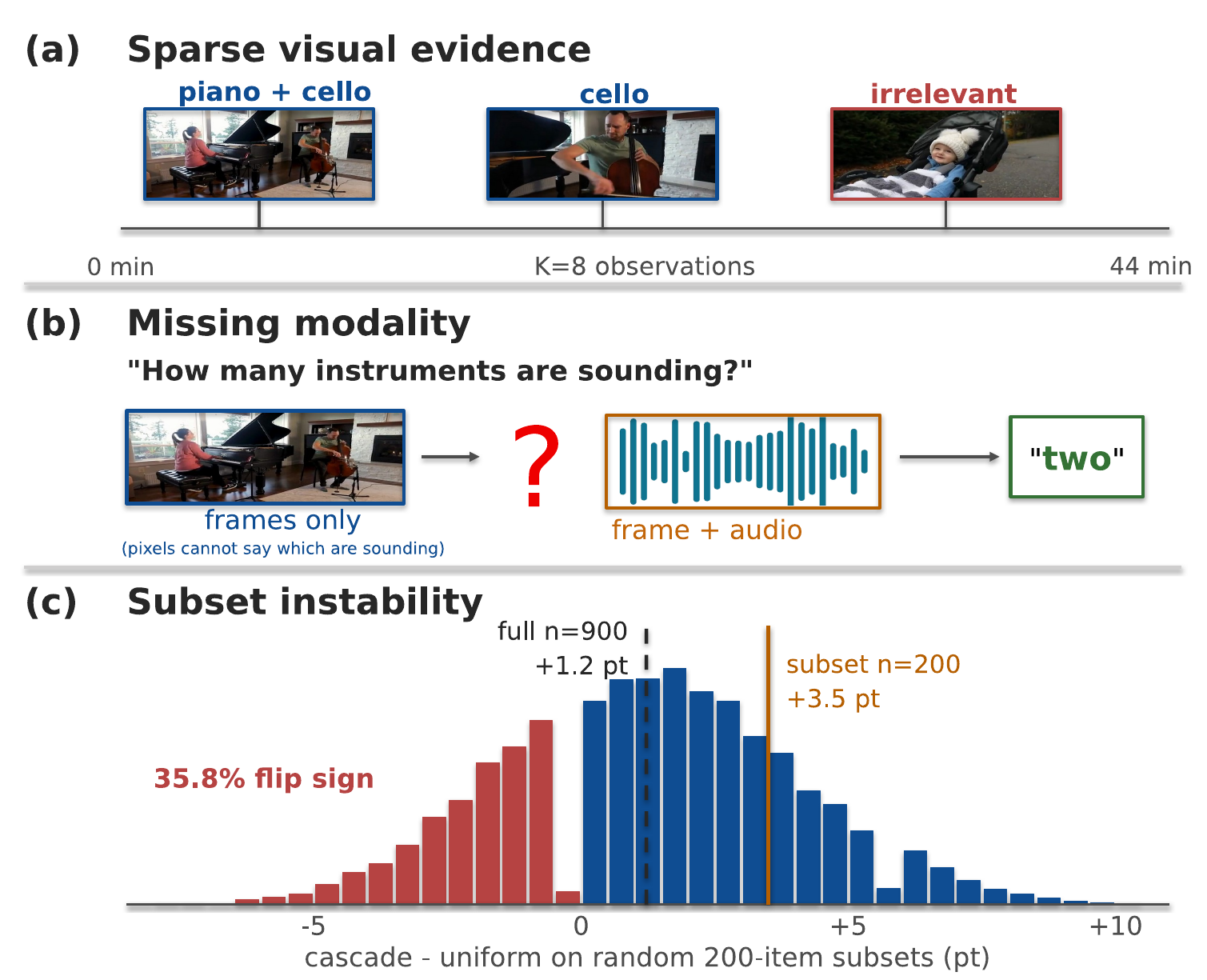}
\vspace{-8pt}
      \caption{\textbf{Three obstacles to long-video QA claims.}
      \textbf{(a)} $K$ frames must cover a 44-minute timeline.
      \textbf{(b)} The real MUSIC-AVQA question needs sound, not pixels alone.
      \textbf{(c)} Across $20{,}000$ random $n{=}200$ subsets of our cached
      $n{=}900$ outcomes, $35.8\%$ reverse the full-bucket effect's sign.}
\label{fig:mainidea}
\end{figure}

Long-video question answering is, at its core, a problem of
\emph{getting the right evidence in front of the model}. A long video
cannot be fed wholesale to a vision--language model (VLM): the standard
pipeline samples a small set of $K$ frames---typically $K{=}8$---and
forwards them. But for \emph{event-sparse} video, where the evidence
that answers a question is concentrated in a handful of frames among
thousands, the question the literature has focused on---\emph{which $K$
frames to pick}---is only half of the problem. Even with the right
frames, an MCQ that asks about narration, on-screen text or off-screen
events can be unanswerable from pixels alone
(Fig.~\ref{fig:mainidea}).

\textbf{Three places to spend a decision.}
A long-video pipeline decides \textbf{(i)} which frames the VLM sees,
\textbf{(ii)} which non-visual signals (Whisper transcript, on-frame
OCR) are injected into its prompt, and \textbf{(iii)} which model
answers. The community has invested heavily in (i)---learned and
submodular selectors, agentic tool calls, tree-structured search---while
(ii) is an implementation detail of caption-augmented baselines and
(iii) is rarely treated as a routing choice. This paper measures all
three under one protocol.

\textbf{An audit, not a leaderboard entry.}
Comparing published long-video QA systems is close to impossible: they
differ simultaneously in reasoner capacity, frame budget, prompt
content, answer-scoring protocol, and evaluation subset. We therefore
introduce \name{}, a single training-free harness in which each axis is an
ablatable stage---a four-stage \emph{propose--filter--align--select}
cascade for the visual axis, an \emph{inject} stage for the textual
axis, and an expert-dispatch step for the model axis---and vary
\emph{one decision at a time} with everything else, including the
grader, held fixed. The audit protocol and its measurements come first,
the reusable harness second; we claim no universally superior selector
or planner.

\textbf{What we find.}
\begin{itemize}\itemsep0pt\topsep2pt
  \item \textbf{(F1) At full Video-MME-long scale, no tested selection
        rule significantly improves on uniform sampling.} On the full
        bucket ($n{=}900$, $K{=}8$) our best selection-only cascade is
        $+1.2$\,pt, $95\%$ CI $[-1.7,+4.1]$, and a $\pm5$\,pt
        equivalence test against uniform excludes gains of that
        size---a bounded effect, not merely an absent one---while a gain
        of a few points is not excluded.
  \item \textbf{(F2) Few-hundred-item subset estimates are unstable.}
        On our $n{=}200$ development prefix three of four contrasts are
        up to $3.8\times$ larger than at $n{=}900$
        (cascade$+$ASR vs.\ uniform $+6.5{\to}+1.7$\,pt) and one
        reverses sign. Resampling $200$-item subsets from the
        full-bucket predictions flips the full-bucket effect's sign in
        $35.8\%$ of draws, and the minimum detectable effect is
        $8.3$--$10.4$\,pt, larger than the $1$--$5$\,pt gains this
        literature reports from a few hundred questions: in our
        setting such subsets cannot establish an effect or certify a
        tight null. We take this as the central result; F1 and F3 support it.
  \item \textbf{(F3) Routing gains are domain-, protocol- and
        model-dependent, and smaller than they first appear.} The pooled
        selection$+$injection gain over uniform on three $n{=}200$ subsets
        ($+4.7$\,pt, stratified McNemar $p{=}0.0017$) is not an injection
        effect: the anomaly-domain ``$+$ASR'' arms injected no text, and
        outside the Video-MME development prefix the pooled contrast is
        $+1.6$\,pt ($p{=}0.24$). With transcripts, ASR appended after the
        question lowers XD-Violence accuracy by $10.0$\,pt, and placing
        the same text before the question removes that harm, though not
        on UCF-Crime, where hallucinated transcripts of silent footage cost
        $13.0$\,pt either way. On Video-MME, text placed first adds
        about $4$\,pt to uniform and cascade frames alike ($p{=}0.002$,
        $0.024$; post hoc, unadjusted). Larger still is \emph{which model answers}:
        dispatching MUSIC-AVQA to an audio-visual expert lifts its worst
        type by $+33.3$\,pt and the micro-average by $+15.1$\,pt
        ($n{=}9{,}129$), yet the same expert with its soundtrack silenced
        keeps $92\%$ of that gain ($69\%$ with another video's
        soundtrack), so this is mostly a model swap, not evidence about audio.
\end{itemize}

\textbf{Contributions.}
\begin{itemize}\itemsep0pt\topsep2pt
  \item \textbf{(C1) \name{}, a matched-conditions audit harness for
        long-video QA.} Backbone, budget $K$, decoding, prompt
        scaffold, grader and items are held fixed while one routing
      decision varies; candidate pools and anchors may differ, so
      pipelines, not rules alone, are compared. The harness, the per-item predictions of every
      reported run and a script that recomputes every statistic in this
      paper offline are released (link in the abstract).
  \item \textbf{(C2) A scale-controlled measurement of three routing
      axes} on Video-MME long, UCF-Crime, XD-Violence (Fighting/Normal),
      MUSIC-AVQA and EgoSchema, at both the subset scale the literature
      uses and full-bucket scale, with paired McNemar tests, bootstrap
      CIs, minimum detectable effects and TOST verdicts
      (Sec.~\ref{sec:experiments}).
  \item \textbf{(C3) An actionable re-allocation of effort.} At $K{=}8$
        on narrative long video the tested selection rules are
        indistinguishable within $\pm5$\,pt, injection effects hinge on
        domain and prompt placement, and the answering model, which
        frame selectors hold fixed, produces the largest contrast. We
        give the effect sizes, costs and failure modes needed to act on
        this (Sec.~\ref{sec:f3},~\ref{sec:discussion}).
\end{itemize}

\textbf{Scope.}
Everything here is \emph{training-free}: off-the-shelf encoders (SigLIP,
Whisper, EasyOCR) and a small instruction-tuned planner LLM
(Qwen2.5-3B-Instruct) reading a one-page markdown memory that was
written after inspecting our development prefix, so planner rows are
diagnostics. Video-MME and EgoSchema use Qwen2.5-VL-7B and the
MUSIC-AVQA base is Qwen2-VL-7B; every table states its backbone.

\begin{figure*}[t]
\centering
\includegraphics[width=\textwidth]{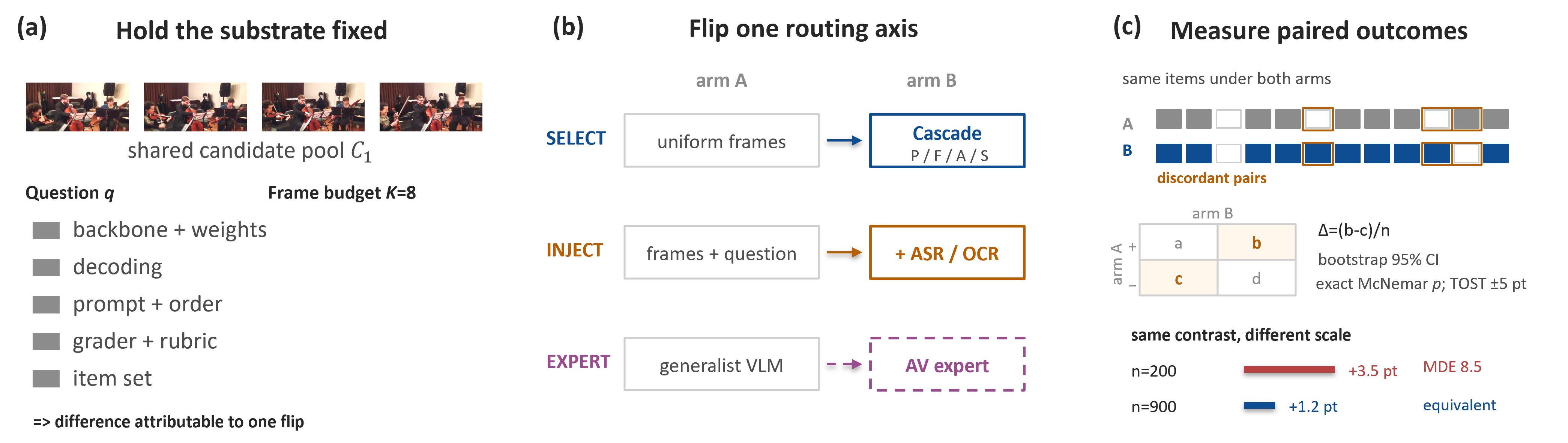}
\vspace{-8pt}
      \caption{\textbf{Matched contrasts are the unit of measurement.}
      \textbf{(a)}~Backbone, budget, decoding, prompt, grader and items
      stay fixed (the ported selectors share one candidate pool; the
      cascade scores its own, Sec.~\ref{sec:setup}). \textbf{(b)}~Exactly one axis flips: SELECT to the cascade,
      INJECT to ASR/OCR, or EXPERT to another model; EXPERT is analyzed
      separately. \textbf{(c)}~Paired outcomes yield $\Delta$, bootstrap CIs,
      exact McNemar $p$ and TOST. See Sec.~\ref{sec:method}.}
\label{fig:system}
\end{figure*}

\section{Related Work}
\label{sec:related}

\textbf{Question-agnostic and single-signal selection.}
Uniform sampling and shot-based selectors ignore the question and remain
the de facto baseline for long-video VLMs; generic video
summarization~\citep{gygli2015subcost} treats the task as diversity-only
subset selection. Learned and language-based variants each commit to one
signal---SeViLA~\citep{yu2023sevila} trains a localizer,
Q-ViD~\citep{romero2023qvid} writes question-conditioned frame captions
for an LLM, LLoVi~\citep{zhang2023llovi} captions densely and then
queries---which is brittle when evidence is localized and invisible to that signal. The
released default proposer is STACFP~\citep{lin2025keyscore}, a
content-adaptive spatio-temporal clustering sampler that self-adapts
between long-take clustering and shot-boundary snapping; the audited runs
predate it (Sec.~\ref{sec:proposefilter}).

\textbf{Caption- and prompt-aware frame scorers.}
KeyScore~\citep{lin2025keyscore} computes a
\emph{contextual drop} signal at CLIP/PE embedding level for
caption- and query-aware frame scoring, and combines it with semantic similarity and
temporal representativeness via a closed-form scalar.
AKS~\citep{tang2025aks} formulates prompt-aware adaptive keyframe
sampling as an optimization of prompt relevance and temporal coverage.
KeyVideoLLM~\citep{liang2024keyvideollm} selects keyframes by
text--frame similarity to compress VideoLLM data. Logic-in-Frames~\citep{guo2025logic}
adds visual--logical dependencies.
The 2025--2026 wave keeps the same embedding-space premise while varying
the \emph{rule}: top-$K$ scoring~\citep{huang2025frag}, temporal
curvature~\citep{miraj2026crest}, wavelet semantic
boundaries~\citep{chen2026wfssb}, prediction
surprise~\citep{kim2026swift}, event anchors~\citep{chen2026efs},
semantic--visual consensus~\citep{sheng2025sevices}, rethinking-based
question-aware selection~\citep{kim2026request}, and entropy-gated budget
scheduling~\citep{li2026ecoframe}. Closest to our cascade,
\citet{zhang2026cses} and \citet{huang2026adaptivegreedy} cast selection
as monotone submodular coverage with a greedy approximation guarantee,
the latter also routing each question to a preset by its type. Our
contribution is not a better rule: at $K{=}8$ on the full Video-MME long
bucket the three rules we ported do not separate from uniform
(Sec.~\ref{sec:f1}); we did not test the rest of this family.

\textbf{Agentic / tool-using video reasoning.}
VideoAgent~\citep{wang2024videoagent}, MoReVQA~\citep{min2024morevqa},
TraveLER~\citep{shang2024traveler}, and
VideoTree~\citep{wang2024videotree} use LLMs to plan,
caption, retrieve and re-localize over several stages or rounds, often
at many model calls per question. These systems are evaluated
end-to-end, so the controller's contribution is not separated from that
of the extra evidence it pulls in. Our planner is much lighter (one
planner call, $\le\!1$\,s, against a fixed tool registry), and we isolate its
contribution with an \texttt{inject\_only} ablation
(Sec.~\ref{sec:procmem}) making the no-inject branch bitwise identical
to the non-agentic baseline---a control we did not find reported in
prior agentic-video work.

\textbf{Audits, and moving past selection.}
Our position follows work that re-measures published gains under
controlled conditions and finds them smaller or absent---in
recommendation~\citep{dacrema2019worrying} and metric
learning~\citep{musgrave2020metric}. Long-video QA is unusually exposed:
evaluation sets are sometimes small, reasoners are expensive enough that few papers re-run
baselines, and open-ended grading adds a protocol degree of freedom.
Closest to us is KFS-Bench~\citep{li2026kfsbench}, which likewise argues
that frame sampling is assessed only \emph{indirectly} through QA
accuracy, and answers with scene-level ground truth. We take the
complementary route: we keep QA accuracy---the quantity the literature
actually claims---and control the protocol around it, then ask whether
reported differences are measurable at the sample sizes used. Concurrent
method-side work also moves beyond one fixed frame set---
VideoRouter~\citep{huang2026videorouter} coordinates global and local
evidence views and GenEvA~\citep{liu2026geneva} aggregates evidence after
selection---or routes by question: VisualRouter~\citep{zhang2026visualrouter}
picks a sampling strategy per query type and HiMu~\citep{benami2026himu}
dispatches predicates to vision, OCR and audio experts, a palette close
to ours. Those works propose mechanisms; we supply a controlled
measurement and extend it to an axis they hold fixed: which model answers.

\textbf{Text-evidence injection and caption-augmented video QA.}
A parallel line gives an LLM text evidence:
LLoVi~\citep{zhang2023llovi} and Q-ViD~\citep{romero2023qvid} feed frame
captions, and the agentic systems above add captioned and retrieved text,
often instead of frames. These systems bundle injection \emph{with}
the rest of their pipeline; we treat it as an ablatable stage in a
controlled selection$\times$injection factorial (Sec.~\ref{sec:f2}) and
show that where the block sits in the prompt changes its effect
(Sec.~\ref{sec:f3}).

\textbf{Procedural memory, MoE, and submodular selection.}
Memory-augmented LLM agents---ReAct~\citep{yao2023react},
Reflexion~\citep{shinn2023reflexion}, Voyager-style skill
libraries~\citep{wang2023voyager}---accumulate experience across
episodes; our \emph{procedural memory} is a one-page markdown of priors
written from outcomes on our own development prefix and appended
verbatim, with no gradient updates. Token-level MoE~\citep{shazeer2017moe} and tool-using
LLMs~\citep{schick2023toolformer} share the gating-over-experts flavour
but operate inside one network or over homogeneous tools, whereas our
expert axis dispatches across separately-trained multimodal models. The
select stage uses lazy-greedy monotone submodular maximization, which is
$(1{-}1/e)$-approximable~\citep{nemhauser1978submodular}.

\textbf{Long-video and audio-visual MLLMs.}
Qwen2-VL~\citep{wang2024qwen2vl} and LLaVA-NeXT-Video are
vision-centric; audio-language
models~\citep{chu2024qwenaudio,deshmukh2023pengi} specialize in acoustic
understanding; audio--visual LLMs such as
VideoLLaMA2-AV~\citep{cheng2024videollama2} integrate both at heavier
cost. Routing across this heterogeneous pool is not addressed by prior
selection methods.

\section{Method}
\label{sec:method}

We are given a video $V$, a question $q$, and a frame budget $K$. A
standard long-video QA pipeline selects a subset
$\mathcal{F}\subseteq V$ of at most $K$ frames and forwards
$(\mathcal{F},q)$ through a fixed vision--language model (VLM)
$e_{\text{VLM}}$. We treat ``what to put in front of the VLM'' as a
richer choice than ``which frames to pick'': non-visual evidence
(Whisper transcript, on-frame OCR) is a second, orthogonal placement
axis. \name{} realizes both as a single training-free pipeline
orchestrated by a lightweight LLM planner reading a procedural memory of
past outcomes.

\subsection{Overview: a Five-Stage, Two-Axis Pipeline}
\label{sec:formulation}
\name{} (Fig.~\ref{fig:system}) progressively narrows thousands of
candidate frames to $K$, then \emph{augments} the resulting visual
context with text evidence before the VLM call:
\[
\begin{aligned}
V \;&\xrightarrow{\text{propose}}\; \mathcal{C}_1
   \;\xrightarrow{\text{filter}}\; \mathcal{C}_2
   \;\xrightarrow{\text{align}}\; (u_q)
   \;\xrightarrow{\text{select}}\; \mathcal{F},\\
(\mathcal{F},q,V) \;&\xrightarrow{\text{inject}}\; (\mathcal{F},\,q\!\oplus\!\mathcal{E})
   \;\xrightarrow{e_{\text{VLM}}}\; \hat{a},
\end{aligned}
\]
with $|\mathcal{F}|\le K$ and $\mathcal{E}$ a (possibly empty) bundle
of text snippets appended to the prompt. Stages 1--4 form the
\emph{visual axis}; stage 5 (\emph{inject}) is the \emph{textual
axis}. A small LLM planner emits a JSON plan that decides, per
question, which proposers, aligners, and injection sources to use.
No component is trained; every selector and inject policy in our
experiments feeds the \emph{same} $e_{\text{VLM}}$ at the \emph{same}
budget $K$, so differences are not due to the reasoner; they reflect the
whole pipeline (candidate pool, filter, anchors), not one rule alone.

\subsection{Skill Library and Planner}
\label{sec:library}
\label{sec:planner}
The cascade draws on reusable skills behind a uniform interface.
\emph{Propose} and \emph{filter} use cheap skills ($O(N)$ per frame, no
GPU encoder): an always-on $\texttt{uniform\_stride}$ backbone, event
audio/visual proposers (EARS~\citep{lin2026ears}), the content-adaptive
$\texttt{stacfp}$ proposer, a motion-difference filter and a
redundancy filter; PySceneDetect shot boundaries are optional.
\emph{Align} skills score each candidate against $q$:
$\texttt{semantic\_siglip}$, $\texttt{object}$, $\texttt{ocr\_text}$,
$\texttt{asr\_whisper}$, $\texttt{clap\_alm}$ and
$\texttt{audio\_energy}$. A single fixed VLM answers. A run's
\emph{planner profile} fixes which cheap skills it enables; every
audited Video-MME, XD-Violence and UCF-Crime run uses the \emph{audit}
profile of Sec.~\ref{sec:proposefilter}.

A planner emits a JSON plan over this fixed registry: proposers and
aligners, fusion weights $\mathbf{w}^{\text{ev}}$, the diversity
bandwidth $\tau$, the uniform-anchor fraction, and the inject sources
$\mathcal{S}^{\text{inj}}\!\subseteq\!\{\texttt{asr},\texttt{ocr}\}$. The
\textbf{rule-based} planner (CPU-bound, the fallback) maps
$(q,\text{dataset id})$ to a plan by keyword matching. The
\textbf{LLM} planner (Qwen2.5-3B-Instruct, one call per question,
$\le\!1$\,s) additionally reads a markdown \emph{procedural memory}
(Sec.~\ref{sec:procmem}) and falls back to the rule-based plan when its
JSON is malformed. The supplementary gives the registry, the keyword
rules, the planner prompt and the schema.

\subsection{Propose and Filter}
\label{sec:proposefilter}
\emph{Propose} unions the enabled proposers' outputs into a candidate
pool $\mathcal{C}_1$. In the \emph{audit} profile a video longer than
$120$\,s gets $66$ event-attention peaks and $66$ evenly spaced
frames ($132$ candidates; $64$ evenly spaced frames for shorter clips).
This proposer fuses audio energy with optical-flow magnitude, but its
motion signal covers only a video's first $4{,}096$ frames, so its peaks
fall in the opening minutes while the uniform half spans the whole video. The
runs were also planned with a PySceneDetect shot-boundary proposer, but
PySceneDetect was not installed, the router skipped the proposer
silently, and we found this only when re-running the pipeline for
release. The released \texttt{audit} profile reproduces the executed
pool (identical selected frames on $24$ of $24$ re-checked items), and
the router now warns whenever a planned tool fails.

\emph{Filter} then keeps the $32$ candidates with the largest grayscale
change from their predecessor in the pool (long videos only). This
frame-difference rule favours high-motion frames and can discard static
evidence such as a motionless speaker or an on-screen sign. The released
default profile instead proposes with STACFP~\citep{lin2025keyscore},
which clusters colour-histogram-plus-timestamp descriptors and adapts
between long takes and edited video, and removes near-duplicate frames
while keeping static-but-distinct ones. At full scale ($n{=}900$)
it scores $46.4\%$, $+0.2$\,pt over uniform ($[-2.7,+3.2]$, equivalent
within $\pm5$\,pt) and $-1.0$\,pt against the audited cascade.

\subsection{Align: Question-Conditioned Utility}
\label{sec:align}
For each candidate $n\in\mathcal{C}_2$, every enabled alignment skill
$s_i$ produces a normalized relevance score
$\tilde\sigma_i(n)\!\in\![0,1]$ against the question. The planner's
fusion weights combine them into a per-frame utility
\[
u_q(n)
\;=\;
\sum_{i\in\mathcal{S}^{\text{ev}}} w^{\text{ev}}_i\,\tilde\sigma_i(n)
\,+\,
\lambda_u\,U(n),
\]
where $U(n)$ is an optional generic informativeness prior (e.g.\ frame
entropy) and $\lambda_u$ a small weight. Whether fusing several cues
here actually produces better frames is one of the questions this paper
measures (Sec.~\ref{sec:f1}).

\subsection{Select: Submodular Maximization}
\label{sec:submodular}
Given the utility $u_q$, we define a temporal coverage kernel
$k_\tau(t,t')=\exp\!\big(-(t-t')^2/2\tau^2\big)$ and the
facility-location set utility
\[
f_q(\mathcal{F})
\;=\;
\sum_{n} \, u_q(n)\,
\max_{m\in\mathcal{F}} k_\tau(t_n,t_m),
\quad |\mathcal{F}|\!\le\!K.
\]
$f_q$ is monotone non-decreasing and submodular; lazy
greedy~\citep{nemhauser1978submodular} returns $\hat{\mathcal{F}}$ with
$f_q(\hat{\mathcal{F}})\ge(1-1/e)\,f_q(\mathcal{F}^*)$. The
hard-indicator special case ($k_\tau\!\in\!\{0,1\}$) recovers temporal
NMS; the Gaussian kernel makes the relevance/diversity trade-off
continuous. Every audited cascade reserves a uniform floor: $\mathrm{round}(K/2)$
of the $K$ frames are evenly spaced anchors that ignore the alignment
scores (Python rounding), so at $K{=}8$ four frames come from
facility-location and four are uniform, while at $K{=}1$ no anchor is
reserved. The floor guarantees baseline temporal coverage when alignment
scores are flat, and it bounds how far any cascade can move from
uniform sampling.

\subsection{Text-Evidence Injection}
\label{sec:inject}
The \emph{inject} stage appends a short text block to the VLM prompt.
For \texttt{asr}, the cached Whisper (tiny.en) transcript chunks within
$\pm w_\text{inj}$ seconds of each selected frame ($w_\text{inj}{=}4$
throughout) become ``[mm:ss]'' lines of at most $240$ characters,
followed by a transcript summary of at most $400$; for \texttt{ocr},
EasyOCR reads each selected frame (at most $160$ characters per frame).
The block follows the question and options, introduced by
``\textit{Additional evidence extracted from the video (treat as
context, not ground truth):}''; placement is a protocol choice that
Sec.~\ref{sec:f3} ablates by moving the identical block before the
question. Over the $n{=}900$ runs ASR injection
adds on average $1{,}344$ characters (median $1{,}320$, $95$th percentile
$1{,}724$; $865$ of $900$ items non-empty) and OCR $274$ ($782$
non-empty). A video with no cached transcript receives no text, and the
harness now reports how many items were injected. Injection is
\emph{ablatable} and matched in \emph{visual budget and decoding
configuration}: the visual context, $K$, the backbone and the decoding
configuration are held fixed across all inject policies, so accuracy
deltas reflect the prompt-side text only. Total context is deliberately
not matched---the injected text is the intervention.

\subsection{Procedural Memory and Inject-Only Isolation}
\label{sec:procmem}
A na\"ive LLM planner may rewrite \emph{every} plan field, conflating
``what cascade do I want?'' with ``should I inject text on this
question?'' We therefore use two constructs. \textbf{Procedural memory}
is a one-page markdown file of priors---which inject sources help on
which question phenotypes (``inject ASR for narration or dialogue;
suppress it for purely visual questions'') and which combinations hurt
(ASR$+$OCR on Video-MME long)---appended verbatim to the planner prompt;
no parameters are updated. Every planner run loads this one file
(reproduced in the supplementary); \texttt{v1} and \texttt{v3} name
planner configurations (free plan; \texttt{inject\_only\_mode}). The file
quotes outcomes on the $n{=}200$ Video-MME evaluation prefix, so planner
numbers are test-set-informed diagnostics, not evidence of
generalisation. \textbf{Inject-only isolation} hard-overrides every plan
field \emph{except} $\mathcal{S}^{\text{inj}}$ with the rule-based
cascade, so only the LLM's per-question inject decision is used and the
no-inject branch is bitwise identical to the no-inject baseline; without
it the LLM silently swapped aligners in $113$ of $200$ plans.

\subsection{Inference}
\label{sec:inference}
Inference runs the five stages of Sec.~\ref{sec:formulation} in order.
The VLM call dominates the cost of the stages that we time; the
planner adds $\le\!1$\,s, and decoding the candidate frames takes about
a third of the VLM call. One-time preprocessing is not in the per-item
times: Whisper transcripts take about $94$\,s per Video-MME long video
and are then cached, so per-question ASR injection is string
concatenation, whereas EasyOCR runs on the selected frames at inference
time ($\approx\!9$\,s per item; plus $\approx\!31$\,s/question end-to-end; supplementary, \emph{Computational Cost}).

\section{Experiments}
\label{sec:experiments}

\subsection{Setup}
\label{sec:setup}
\textbf{Benchmarks.} We evaluate the selection and injection axes on
three long-video evaluation domains spanning the narrative--anomaly
axis: \textbf{Video-MME
long}~\cite{fu2024videomme} (full bucket, $n{=}900$; plus the $n{=}200$
subset used for the paired factorial ablations),
\textbf{UCF-Crime}~\cite{sultani2018ucfcrime} and
\textbf{XD-Violence}~\cite{wu2020xdviolence} ($n{=}200$ balanced
binary event-QA each; XD: Fighting/Normal). We use
\textbf{MUSIC-AVQA}~\cite{li2022musicavqa}
for the expert axis and EgoSchema~\cite{mangalam2023egoschema} ($n{=}500$)
as a cross-benchmark check (supplementary, \emph{Extended Experiments}).

\textbf{What is held fixed.} Every contrast is paired and matched: the
same questions, backbone, frame budget $K$ (unless noted, $K{=}8$),
prompt scaffold, greedy decoding and grader; only the routing decision
under test varies. The three ported selectors score one $256$-frame pool
strided over the whole video, whereas our cascade scores at most $32$
keyframes from its own proposal pool (Sec.~\ref{sec:proposefilter}) and
keeps half of its frames as uniform anchors, so the pools differ by
design. Uniform-$K$ uses endpoint-inclusive evenly spaced frames (at
$K{=}1$ the video's first frame); a segment-midpoint variant is checked
at $K{=}8$ and $K{=}1$. We port the \emph{selection rule} of AKS, KeyScore and
VideoTree, not the published systems (their candidate generation,
backbone and prompt are replaced); the KeyScore port reproduces the
reference scores, and because the AKS budget $\lfloor K/2^d\rfloor$
vanishes at depth $d{\ge}4$ for $K{=}8$, its port is a binned top-$K$
(supplementary, \emph{Selector ports}). These rows bound the selection
component, not the end-to-end pipelines. Video-MME and EgoSchema use
Qwen2.5-VL-7B-Instruct and MUSIC-AVQA Qwen2-VL-7B-Instruct; a
cross-backbone check is in the supplementary.

\textbf{Statistics.} Pairwise significance is two-sided exact McNemar on
per-item correctness. Headline contrasts also get a $95\%$ paired
bootstrap CI ($10{,}000$ resamples), the discordant-pair count with its
minimum detectable effect (MDE) and, where a null is claimed, a TOST
verdict on the $90\%$ CI. The $\pm5$\,pt margin is the top of the range
of selection-axis gains this literature claims ($1$--$5$\,pt); it is a
choice, not a noise floor, so ``equivalent'' means only that gains of
that size are excluded. We also report $\pm3$\,pt, which no full-bucket
selection contrast passes. Video-MME long asks three questions per
video, so every full-bucket interval was recomputed with a
video-clustered bootstrap: its standard error is at most $1.02\times$
the item-level one and no verdict changes (supplementary). F1's
full-bucket contrasts and F3's expert contrast are confirmatory; the
cells of Table~\ref{tab:f3domain}, the pooled F3(a) contrast
(confirmatory only outside the $200$-item development prefix), the
placement contrasts, the budget sweep and the supplementary sweeps are
exploratory, with unadjusted $p$-values. A $p{>}0.05$ is never described
as a win.

\textbf{Grading.} Multiple-choice benchmarks (Video-MME, EgoSchema,
UCF-Crime, XD-Violence) are scored by exact match of the parsed option
letter. Open-ended answers (MUSIC-AVQA) are graded by a greedy LLM judge
(Qwen2.5-VL-32B text path, the standard GPT-assisted prompt of
VideoLLaMA~2). On $150$ stratified decisions it matches a blind human
re-annotation at $94.0\%$ ($\kappa{=}0.88$): $100\%$ ($90/90$; $95\%$
Wilson interval $[95.9,100]$) on MUSIC-AVQA, which carries
Sec.~\ref{sec:f3}, and $85.0\%$ ($51/60$) on EgoTextVQA, which supports
no claim here (supplementary, \emph{Judge Validation}).

\subsection{F1 --- At Full Scale, No Tested Selection Rule Separates from
Uniform Sampling}
\label{sec:f1}
Table~\ref{tab:scale900} is the paper's primary result: all methods
evaluated on the \emph{entire} Video-MME long bucket, paired across
methods. \emph{No tested} selection rule significantly improves on uniform
sampling. The largest selection-only effect is $+1.9$\,pt (ported AKS,
$p{=}0.273$) and our cascade is $+1.2$\,pt ($95\%$ CI $[-1.7,+4.1]$);
both pass a TOST equivalence test against uniform at $\pm5$\,pt (TOST
uses the $90\%$ CI, which is why AKS passes despite a $95\%$ upper bound
of $+5.0$)---a bounded effect, not merely an absent one. The bound is not
arbitrarily tight: at $\pm3$\,pt \emph{no} selection-only row passes
(cascade $90\%$ CI $[-1.2,+3.7]$, AKS $[-0.8,+4.6]$), so the full bucket
rules out a literature-scale ($\ge\!5$\,pt) selection gain while leaving
a small one unresolved. A segment-midpoint uniform baseline scores the
same, and the cascade and AKS stay equivalent to it. The rows below $p{=}0.05$ are not selection
rules but post hoc injection variants: the ASR block placed first, with
cascade or uniform frames alike (Sec.~\ref{sec:f3}).

F1 says that at $K{=}8$ on dense narrative long video the tested
selection rules produce near-equivalent accuracy on this backbone. We can
bound the effect but not attribute it: \emph{coverage saturation}
(uniform sampling may already place enough answer-relevant evidence in
the prompt) and \emph{reasoner insensitivity} (the VLM does not reliably
exploit better frames) both fit the data. Two design facts also bound how
far the cascade can move from uniform: half of its frames are uniform
anchors and its pool has $132$ candidates (Sec.~\ref{sec:proposefilter});
the ported selectors score a $256$-frame pool and land in the same
place, but at $K{=}8$ the official AKS rule returns no frame, so its port
is a binned top-$K$ (supplementary, \emph{Selector ports}). F1 does
\emph{not} say selection never matters: the cascade's point estimates on
the two anomaly domains ($+4.0$ and $+3.5$\,pt at $n{=}200$, neither
significant) remain hypotheses. At $K{=}1$ the $n{=}200$ gain
($+10.0$\,pt, $p{=}0.009$) shrinks to $+2.7$\,pt at $n{=}900$
($[-0.7,+6.0]$, $p{=}0.13$), and a middle-frame uniform baseline beats
uniform's first frame by $+5.1$\,pt ($p{=}0.003$) and the cascade by
$2.4$\,pt (n.s.): that effect was a weak baseline. On
EgoSchema ($n{=}500$) the cascade is $-1.6$\,pt ($[-4.6,+1.4]$) and the
AKS, KeyScore and VideoTree ports $-2.6$, $-4.0$ and $-0.2$\,pt: none helps.

\begin{table}[t]
\centering
\caption{\textbf{F1: the full-scale audit.} All methods on the complete
Video-MME long bucket ($n{=}900$, $K{=}8$, Qwen2.5-VL-7B), paired across
methods; $\Delta$ and CI are against uniform and $p$ is exact McNemar,
unadjusted. \emph{No tested selection rule significantly improves on
uniform}, and every selection-only row is equivalent to it within
$\pm5$\,pt (none within $\pm3$). $^\dagger$Post hoc placement variant of
ASR injection (Sec.~\ref{sec:f3}); the only rows with $p{<}0.05$. Wall:
mean \texttt{elapsed\_sec}/$n$ on one GB10, shared with other jobs,
excluding one-time Whisper ($\approx\!31$\,s/question) and including
EasyOCR.}
\label{tab:scale900}
\small
\resizebox{\columnwidth}{!}{%
\begin{tabular}{lccccc}
\toprule
Method ($K{=}8$, $n{=}900$) & Acc.\ (\%) & $\Delta$ & 95\% CI & $p$ & Wall s/item \\
\midrule
Uniform-$K$                        & 46.2 & ---    & ---            & ---   & 10.0 \\
Midpoint uniform-$K$               & 46.2 & $0.0$  & $[-2.8,+2.9]$  & 1.000 & --- \\
AKS~\cite{tang2025aks} (port)      & 48.1 & $+1.9$ & $[-1.2,+5.0]$  & 0.273 & 22.3 \\
KeyScore~\cite{lin2025keyscore} (port) & 47.1 & $+0.9$ & $[-2.2,+4.0]$ & 0.624 & 22.4 \\
VideoTree~\cite{wang2024videotree} (port) & 47.8 & $+1.6$ & $[-1.6,+4.8]$ & 0.360 & 22.3 \\
Cascade (selection only)           & 47.4 & $+1.2$ & $[-1.7,+4.1]$  & 0.452 & 13.3 \\
Cascade $+$ OCR                    & 47.2 & $+1.0$ & $[-1.9,+4.0]$  & 0.552 & 21.6 \\
Cascade $+$ ASR                    & 47.9 & $+1.7$ & $[-1.6,+4.8]$  & 0.337 & 12.8 \\
Cascade $+$ ASR, text first$^\dagger$ & 50.2 & $+4.0$ & $[+0.6,+7.3]$  & 0.024 & --- \\
Uniform $+$ ASR, text first$^\dagger$ & 50.7 & $+4.4$ & $[+1.8,+7.2]$  & 0.002 & --- \\
ProcMem planner                    & 48.4 & $+2.2$ & $[-0.7,+5.1]$ & 0.154 & 15.7 \\
\bottomrule
\end{tabular}}
\end{table}

\subsection{F2 --- Subset-Scale Effect Estimates Are Unstable}
\label{sec:f2}
\begin{figure}[t]
\centering
\includegraphics[width=\columnwidth]{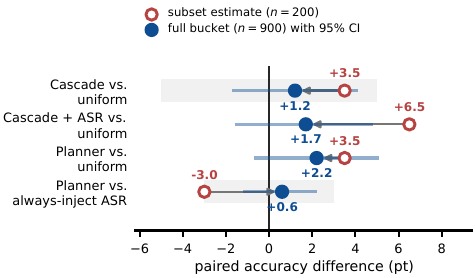}
\caption{\textbf{Subset-scale effect estimates are unstable.}
Matched contrasts at $n{=}200$ and $n{=}900$.
Segments show full-bucket paired-bootstrap $95\%$ CIs; grey bands mark
the equivalence margin of the contrast ($\pm3$\,pt for planner vs.\
always-inject ASR, $\pm5$\,pt for cascade vs.\ uniform). One contrast
reverses sign.}
\label{fig:scalegap}
\end{figure}

Component-level contrasts in this area are often measured on a few
hundred questions (supplementary, \emph{Where the $1$--$5$\,pt range comes from}), and our own
development prefix has $200$. Fig.~\ref{fig:scalegap} places our
subset-scale measurements next to the full-bucket measurement of the \emph{same}
contrast, on the \emph{same} pipeline, with the \emph{same} grader.
Three of the four subset estimates exceed their full-bucket
counterparts---by $1.6$--$3.8\times$---and the fourth
(\emph{planner vs.\ always-inject ASR}) reverses: $-3.0$\,pt at
$n{=}200$, $+0.6$\,pt at $n{=}900$. Had we reported only the subset, we
would have drawn the opposite conclusion about our own planner.

\textbf{The gap is sampling variability, not an unlucky subset.} Our
$n{=}200$ set is a deterministic \emph{prefix} of the $n{=}900$ set (the
first $67$ of $300$ videos), so the full-bucket predictions already
contain it and we can ask, with no new inference, what a differently drawn
subset would have shown. Over $B{=}20{,}000$ random $200$-item subsets our
prefix is unremarkable ($+3.5$\,pt, $82$nd percentile); the width is the
finding: a $10.5$\,pt $95\%$ range, a sign flip of the full-bucket effect
in $35.8\%$ of draws, and $p{<}0.05$ in only $2.8\%$. Drawing \emph{videos}
rather than items (three questions per video; $67$ of $300$ clusters)
gives $[-4.0,+6.5]$ and $35.7\%$, so within-video correlation does not
explain it. For cascade$+$ASR ($+6.5{\to}+1.7$\,pt) the range is
$[-4.5,+7.5]$ with $31.9\%$ sign flips (supplementary).

\textbf{The arithmetic behind it.} McNemar's test uses only the
\emph{discordant} items: cascade vs.\ uniform at $n{=}200$ has $37$
($b{=}22$, $c{=}15$) and a minimum detectable effect (MDE) of $8.5$\,pt
at $80\%$ power; across the four selection-axis contrasts the MDE is
$8.3$--$10.4$\,pt (supplementary). Such a comparison can neither
establish a few-point effect nor certify a tight null, so such claims
need full-bucket scale or a reported MDE.

\subsection{F3 --- Routing Gains Depend on Domain and Answering Model}
\label{sec:f3}
If neither selection nor injection separates at full scale on Video-MME,
what does? Two things, both different from what the subset-scale numbers
suggest.

\textbf{(a) Injection depends on the domain and on where the text sits in
the prompt.} Our protocol appends the evidence block after the question,
options and answer instruction. On narrative video, ASR so appended adds
$+3.0$\,pt within the cascade on the $200$-item development prefix
($p{=}0.18$) and $+0.4$\,pt on the full bucket ($[-1.6,+2.4]$,
$p{=}0.75$): no effect is established. On the anomaly clips the published
``$+$ASR'' arms injected nothing: no transcript had been computed for
them, so their $+4.0$ and $+3.5$\,pt over uniform are selection effects
of the plain cascade (Table~\ref{tab:f3domain}). With Whisper transcripts
built for all $400$ clips, the appended block \emph{lowers} XD-Violence
accuracy from $91.0$ to $81.0\%$ ($-10.0$\,pt, $95\%$ CI $[-14.5,-5.5]$,
$b/c{=}2/22$, $p{<}10^{-4}$) and UCF-Crime from $85.0$ to $72.0\%$
($-13.0$\,pt, $[-18.0,-8.5]$, $b/c{=}1/27$). On XD the harm is a
placement effect: the identical text placed \emph{before} the question
leaves accuracy at $91.0\%$, the cascade's ($0.0$\,pt, $[-3.0,+3.0]$;
$+10.0$\,pt over appending, $p{=}0.0003$). On UCF-Crime it is not: placing
the text first gives the same $72.0\%$, because $160$ of the $200$
transcripts of that silent surveillance footage contain Whisper's
hallucinated ``thank you for watching''. On Video-MME the text placed
first lifts the cascade to $50.2\%$ ($+2.8$\,pt, $p{=}0.055$) and uniform
sampling to $50.7\%$ ($+4.4$\,pt, $[+1.8,+7.2]$, $p{=}0.002$), the two
being indistinguishable ($p{=}0.83$), where appending gave $+0.4$\,pt: the
gain follows the text and its position, not the frames. With uniform frames,
text between the question and the options, or between the options and the
instruction, gives only $+1.0$ and $+1.9$\,pt (n.s.), like appending
($+1.6$); only text before the question gives $+4.4$ (supplement). On
Qwen2-VL-7B it replicates: $52.7\%$ text first ($+7.2$\,pt, $p{<}10^{-6}$),
$47.2\%$ appended ($+1.8$, $p{=}0.09$). Placement was varied \emph{after}
the anomaly results (unadjusted); injection effects are protocol- and
transcript-sensitive in sign, not just size.

The published pooled estimate was not an injection effect either.
Over the three $n{=}200$ subsets cascade$+$ASR beat uniform by $+4.7$\,pt
($95\%$ CI $[+1.8,+7.7]$; stratified McNemar $p{=}0.0017$), but the
anomaly arms were the plain cascade, and the first $200$ Video-MME items
are the prefix on which the cascade and planner memory were developed.
Outside that prefix ($n{=}1{,}100$) the contrast is $+1.6$\,pt
($p{=}0.24$), and with transcripts in the anomaly arms $-3.0$\,pt
($[-6.2,0.0]$). The planner never injected on the anomaly clips, so its
$\pm2$\,pt differences there lie inside the noise band F2 quantifies.

\begin{figure}[t]
\centering
\includegraphics[width=\columnwidth]{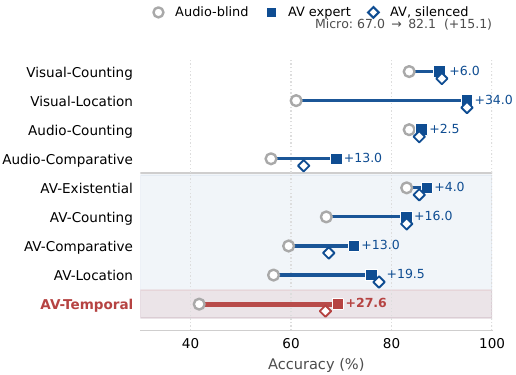}
\caption{\textbf{Answering-model choice produces the largest measured
contrast, and little of it is audio.} MUSIC-AVQA, $K{=}8$, same judge,
the same $200$ questions per type: audio-blind Qwen2-VL-7B (circle), the
audio-visual expert (square) and the same expert with its soundtrack
silenced (diamond). Labels give the swap (circle to square).}
\label{fig:avexpert}
\end{figure}

\textbf{(b) Changing which model answers yields the largest measured
contrast, but little of it is audio.}
On the \emph{full} MUSIC-AVQA split ($n{=}9{,}129$, same questions, both
models at $K{=}8$, same judge) a generalist visual VLM (Qwen2-VL-7B)
reaches $37.6\%$ on \textbf{AV-Temporal}---``which instrument sounds
\emph{first}''---while the audio-visual expert
VideoLLaMA2-AV~\cite{cheng2024videollama2} reaches $70.9\%$ ($+33.3$\,pt,
$95\%$ CI $[+29.6,+37.1]$, exact McNemar $p{=}2.8\!\times\!10^{-54}$), and
the micro-average moves from $66.1$ to $81.2\%$. \emph{Every} type
improves, but this is a \emph{model swap}: purely visual Visual-Location
moves $+27.3$\,pt, and the expert's audio-visual training mix includes
$\approx\!35$k AVQA-music (MUSIC-AVQA) entries~\cite{cheng2024videollama2}.
We therefore re-ran the \emph{same} expert with its soundtrack replaced
by digital silence on $200$ questions per type ($1{,}799$ usable), all
three arms on the same questions and graded by the same judge
(Fig.~\ref{fig:avexpert}). Post-stratified to the full split's type mix,
the silenced expert scores $80.9\%$, against $82.1\%$ with sound and
$67.0\%$ for the audio-blind model: silencing removes $1.3$\,pt
($95\%$ CI $[-0.1,+2.6]$) of the $+15.1$\,pt swap, so $92\%$ of the gain
comes with the model. Giving each video another video's soundtrack scores
$77.5\%$: the expert does read the audio (the wrong soundtrack costs
$3.4$\,pt against silence, $p{=}10^{-5}$, and the right one adds $4.7$ over
it), yet it still beats the audio-blind model by $+10.5$\,pt ($69\%$ of
the swap). The contrast is mostly a model swap with a smaller audio part.

\textbf{Question-only dispatch.} A fixed audio/speech keyword rule picks
the expert from the question text alone; replayed on both passes it reaches $78.6\%$, $+12.5$\,pt over the visual base ($95\%$ CI
$[+11.5,+13.5]$) but $2.6$\,pt below always using the AV expert, to which
it sends $84.9\%$ of questions: on this modality-homogeneous benchmark
routing collapses to that expert, an accuracy gain, not a cost saving.

\begin{table}[t]
\centering
\caption{\textbf{F3(a): the sign of an ASR effect depends on the domain and
on placement.} Qwen2.5-VL-7B, $K{=}8$, $n{=}200$ per domain (Video-MME: first
$200$ items). \emph{After}: block appended after the question and options
(audited protocol); \emph{before}: ahead of the question.
$^\dagger$No transcript existed: nothing injected. Planner row: domain-specific
configurations (never injected on anomaly clips). Cells are exploratory.}
\label{tab:f3domain}
\small
\resizebox{\columnwidth}{!}{%
\begin{tabular}{lccc}
\toprule
Method ($K{=}8$, $n{=}200$)        & Video-MME long & XD F/N & UCF-Crime \\
\midrule
Uniform-$K$ (no inject)             & 50.0           & 87.0        & 81.5 \\
Cascade (no inject)                 & 53.5           & 91.0        & 85.0 \\
Cascade $+$ ASR, as published       & 56.5  & 91.0$^\dagger$ & 85.0$^\dagger$ \\
Cascade $+$ ASR, after the question & 56.5  & 81.0 & 72.0 \\
Cascade $+$ ASR, before the question & 57.0 & 91.0 & 72.0 \\
Cascade $+$ always-inject OCR       & 55.5           & 89.5        & 83.0 \\
ProcMem planner                     & 53.5           & 92.5 & 83.0 \\
\bottomrule
\end{tabular}}
\end{table}

\section{Discussion and Limitations}
\label{sec:discussion}

\textbf{Interpreting the audit.} Coverage saturation and reasoner
insensitivity both fit F1; we do not separate them. F2 concerns the
precision of our controlled contrasts, not a re-run of prior systems
under their original protocols. F3's accuracy headroom from the answering
model does not establish cheaper inference.

\textbf{Limitations.}
(i)~The expert contrast is a \emph{model swap}: VideoLLaMA2-AV and
Qwen2-VL-7B differ in far more than audio access, the expert trained on
MUSIC-AVQA data, and the audio controls (silenced, mismatched) cover a
type-stratified subset, one expert, one random soundtrack assignment.
(ii)~F1 is established on Video-MME long at $K{=}8$, $n{=}900$, with
Qwen2.5-VL-7B; equivalence holds at $\pm5$ but not $\pm3$\,pt.
(iii)~The audited cascade is one configuration (event-attention and
uniform proposals, a motion filter, $50\%$ uniform anchors;
Sec.~\ref{sec:proposefilter}); the released default profile differs. The ported selectors reimplement selection rules only; results do not bear on the originals.
(iv)~The planner's memory was written after inspecting the $n{=}200$
development prefix, so planner numbers are test-set-informed, and
Table~\ref{tab:f3domain} and the placement variants are post hoc and unadjusted; the anomaly-QA sets are binary conversions; generalization beyond these datasets, models and $K$ is untested.

\section{Conclusion}
\label{sec:conclusion}
We audit long-video QA by separating frame selection, evidence injection
and answering-model choice. At $K{=}8$ on full Video-MME long, no tested
selector improves significantly over uniform (equivalent within $\pm5$,
not $\pm3$\,pt), and subset estimates are unstable. The audit also
overturned our own first results: the anomaly-domain ``$+$ASR'' arms
injected nothing, the pooled gain shrinks outside the development prefix,
and an injection effect depends on where the text sits in the prompt.
The largest expert contrast is mostly a model swap ($92\%$ survives a
silenced soundtrack, $69\%$ a mismatched one), and question-only dispatch
recovers accuracy without a demonstrated saving.

The next step is not simply to select better frames, inject more
evidence or route to stronger models, but to identify which decision
limits performance, measure its contribution under matched conditions,
and add complexity only where the gain is reliable.

\clearpage

\clearpage
\begin{center}
{\large\bfseries Supplementary Material}
\end{center}
\appendix

\section{Procedural Memory: Extended Analysis}
\label{app:procmem}

\subsection*{What the memory is (and is not)}
The procedural memory is a single markdown file of $2{,}028$ characters
(\texttt{configs/procedural\_memory\_v1.md}, reproduced below) appended
verbatim to the planner LLM's system prompt before the few-shot examples.
It is not a learned parameter, an embedding store or a retrieval index:
the LLM reads the text anew for each question. \emph{Every} reported
planner run loads this one file. The labels ``v1'', ``v2'' and ``v3'' in
the main paper and in our run names denote planner \emph{configurations},
not different memory texts: v1 lets the LLM rewrite the whole plan, v2
freezes the aligner mix, and v3 is \texttt{inject\_only\_mode}. A second
file, \texttt{procedural\_memory\_v4.md}, was drafted but no reported run
uses it. The memory text is:
\begin{quote}\small\raggedright
\emph{Past empirical findings (HIGH PRIORITY---override your default
conservatism).}
(1)~Long documentary or educational videos (duration $>$120\,s) almost
always contain narration that carries the answer even when the question
does not mention a narrator; for ``what is the video about'', ``what
does the speaker explain'', ``why'' or ``how'' questions \emph{default
to} \texttt{inject\_text=["asr"]}. ``In our logs this gave $+6.5$\,pt
accuracy ($p{=}0.066$) versus no injection. Skipping injection lost
$4$\,pt.''
(2)~For purely visual attribute questions (colour, count, position,
clothing, who holds what), ASR adds noise: use \texttt{[]}.
(3)~Never inject ASR and OCR together: ``the combo ASR+OCR lost $1.5$\,pt
versus ASR alone ($0.550$ vs $0.565$)''; pick at most one stream.
(4)~For on-screen-text questions use \texttt{["ocr"]} only.
(5)~For short ($\le$120\,s) anomaly clips, OCR of captions sometimes
helps and ASR is usually too sparse; prefer \texttt{["ocr"]} when in
doubt.
\emph{Decision shortcut for long videos:} visual-attribute question
$\to$ \texttt{[]}; on-screen text $\to$ \texttt{["ocr"]}; otherwise
$\to$ \texttt{["asr"]}.
\end{quote}
The file header states that it was mined from Video-MME long execution
logs at $n{=}200$, $K{=}8$ with Qwen2.5-VL-7B, and the quoted figures
($+6.5$\,pt, $0.550$ vs $0.565$) are outcomes on the same first $200$
items that the $n{=}900$ evaluation contains.

\subsection*{The single controlled degree of freedom}
To attribute accuracy changes to the LLM's routing decision cleanly, we
deploy \texttt{inject\_only\_mode}: every plan field \emph{except}
per-question \texttt{inject\_text} is hard-overridden by the rule-based
cascade. This ensures the no-inject branch of the LLM planner is bitwise
identical to the no-inject cascade baseline. Without this isolation, the
LLM was observed to silently swap aligners ($113$ of $200$ plans on
Video-MME long switched \texttt{semantic\_siglip} to
\texttt{object\_siglip}), which would contaminate the attribution to the
planner's inject decision.

\paragraph{Evaluation independence.}
The memory is test-set-informed: it was written after inspecting
outcomes on the $n{=}200$ Video-MME evaluation prefix, and its text
quotes them. The v2 and v3 configurations were also revised after
inspecting that prefix. The $n{=}900$ planner row therefore mixes $200$
development items with $700$ items that neither the memory nor the
configuration saw. On those $700$ items the planner scores $47.0\%$
against $45.4\%$ for always-inject ASR ($+1.6$\,pt, exact $p{=}0.17$)
and $45.1\%$ for uniform ($+1.9$\,pt, $p{=}0.31$); on the $200$
development items it scores $53.5\%$ against $56.5\%$ for always-inject
ASR. Neither held-out difference is significant, so planner accuracy is
not evidence of generalisation, and the matched selector and injection
contrasts carry the audit findings.

\subsection*{When the planner pays for itself}
The planner makes one call per question ($\le 1$ second), but is not
established as better than always-inject ASR: on the $n{=}200$ prefix
the differences are $-3.0$\,pt on Video-MME, $+1.5$\,pt on the
XD-Violence Fighting/Normal subset and $-2.0$\,pt on UCF-Crime, with
neither anomaly-domain contrast significant. On the anomaly clips the
planner never chose to inject (none of the $400$ questions), so those
differences reflect the tools the LLM chose, not its injection
decisions; no ASR transcript existed for those clips either
(Sec.~\ref{app:anomasr}). At $n{=}900$ the planner is
numerically strongest ($48.4\%$) but equivalent to always-inject ASR
within $\pm3$\,pt. The per-item decisions of every planner run are
released with the predictions, so the planner row can be recomputed
without calling the LLM.

\section{Judge Validation Data}
\label{app:judge}

\paragraph{Where a judge is used.} Video-MME, EgoSchema, UCF-Crime and
XD-Violence are multiple choice: the answering model's reply is parsed to
an option letter and compared to the key by exact match, with no judge.
Only open-ended answers (MUSIC-AVQA, and EgoTextVQA in an auxiliary
study) are graded by the judge: Qwen2.5-VL-32B-Instruct used through its
text path, greedy decoding, with the GPT-assisted evaluation prompt of
VideoLLaMA~2 and EgoTextVQA (question, correct answer and predicted
answer in; ``yes''/``no'' and a $0$--$5$ score out).

\paragraph{Protocol and result.} One author blindly re-annotated $150$
stratified accept/reject decisions without seeing the judge's labels.
Agreement was $94.0\%$ with Cohen's $\kappa{=}0.88$ ($95\%$ bootstrap CI
$[0.80,0.95]$). It differs sharply by dataset:
\begin{center}\small
\begin{tabular}{lcccc}
\toprule
Subset & $n$ & Agree & FA/FR & $\kappa$ \\
\midrule
MUSIC-AVQA  & 90  & $100\%$ [95.9,100] & 0/0 & 1.00 \\
EgoTextVQA  & 60  & $85.0\%$ [73.9,91.9] & 2/7 & 0.70 \\
All         & 150 & $94.0\%$ [89.0,96.8] & 2/7 & 0.88 \\
\bottomrule
\end{tabular}
\end{center}
(brackets: $95\%$ Wilson intervals; FA/FR: false accepts / false
rejects.) All nine disagreements are on EgoTextVQA, which supports no
headline claim; the $90$ MUSIC-AVQA decisions behind the expert-axis
analysis all agreed. The judge errs conservatively where it errs
($7$ false rejects vs.\ $2$ false accepts). These annotations validate
grader behavior only; they are not benchmark results. The per-decision
table is released (\texttt{results/judge\_validation/judge\_vs\_human.csv})
and \texttt{scripts/judge\_validation\_report.py} recomputes the table.

\section{Cross-Backbone Robustness}
\label{app:xvlm}

To check that F1 and F2 are not specific to one reasoner we repeat the
Video-MME long contrasts on a second vision--language backbone,
Qwen2-VL-7B-Instruct, through the identical cascade and injection code
path at full scale ($n{=}900$, $K{=}8$).

\begin{table}[h]
\centering
\caption{\textbf{Second backbone, full bucket.} Qwen2-VL-7B-Instruct on
Video-MME long ($n{=}900$, $K{=}8$); $\Delta$ is against uniform with its
$95\%$ paired-bootstrap interval, $p$ is exact McNemar. $^\ast$The original
$+$ASR arm was executed after PySceneDetect had been installed, so it scored
$198$ candidates against $132$ for the other arms; the pool-matched re-run
uses the audited profile like the cascade.}
\label{tab:xvlm}
\small
\setlength{\tabcolsep}{4pt}
\resizebox{\columnwidth}{!}{%
\begin{tabular}{lcccc}
\toprule
Arm & Pool & Acc.\ (\%) & $\Delta$ & $p$ \\
\midrule
Uniform-$K$                  & ---  & 45.4 & ---                    & --- \\
Cascade                      & 132  & 44.8 & $-0.7$ $[-3.4,+2.0]$   & 0.685 \\
Cascade $+$ ASR$^\ast$ (original) & 198 & 48.0 & $+2.6$ $[-0.4,+5.6]$ & 0.107 \\
Cascade $+$ ASR (pool-matched) & 132 & 44.6 & $-0.9$ $[-3.8,+2.1]$ & 0.614 \\
\bottomrule
\end{tabular}}
\end{table}

The selection-only contrast replicates the F1 null on the second
backbone: the cascade is $-0.7$\,pt below uniform ($b/c{=}73/79$,
$90\%$ CI $[-2.9,+1.6]$), equivalent within $\pm5$\,pt. The original
$+$ASR arm was not comparable: its pool differed, and its $+3.2$\,pt gain
over the cascade ($p{=}0.007$) disappears when the arm is re-run with the
same $132$-candidate pool ($44.6\%$; $-0.2$\,pt within the cascade,
$[-2.3,+1.9]$, $p{=}0.92$; $-0.9$\,pt against uniform, $p{=}0.61$), so that
gain was the candidate pool, not the injected text. On the first $200$ items the same runs score $45.0$, $49.5$
and $50.5\%$ (uniform, cascade, cascade$+$ASR): the cascade's $+4.5$\,pt
on the development prefix becomes $-0.7$\,pt on the full bucket, the F2
pattern on a second backbone. The $n{=}200$ planner arm ($45.0\%$) ran with
a different candidate pool from its comparison arms (the STACFP profile,
$197$ candidates), so we make no claim that the planner does or does not
transfer across backbones.

\section{Construction of the Anomaly-QA Evaluations}
\label{app:qaconstruction}

UCF-Crime and XD-Violence are anomaly-recognition datasets rather than
native QA benchmarks. We convert them deterministically to binary QA;
the construction below defines the exact evaluated subsets.

\paragraph{UCF-Crime.}
We read the official \texttt{Anomaly\_Test.txt} split. Of 290 resolvable
videos, 140 are anomalous and 150 normal. We alternate the two groups
in their native split order and take the first 200, yielding 100
anomalous and 100 normal videos. The anomalous subset contains Abuse
(2), Arrest (5), Arson (9), Assault (3), Burglary (13), Explosion (21),
Fighting (5), RoadAccidents (23), Robbery (5), and Shooting (14).
Every video receives one question: ``Does this video contain any
unusual, dangerous, or criminal activity?'' with choices
\{A: Yes, B: No\}; every anomaly maps to A and every normal video to B.

\paragraph{XD-Violence.}
The local test split contains 120 Fighting, 84 Shooting, and 300 Normal
videos. Files are sorted within folders ordered Fighting, Shooting,
Normal; positive and normal lists are then alternated before truncation.
Consequently, the reported $n{=}200$ deterministic prefix contains 100
Fighting and 100 Normal videos and no Shooting videos. Each receives
``Does this video contain violent activity such as fighting or
shooting?'' with \{A: Yes, B: No\}; Fighting/Shooting map to A and
Normal to B. We therefore interpret this result as a
Fighting-vs.-Normal XD-Violence subset, not full category coverage.

\paragraph{Common controls.}
Both conversions use one item per video, exact A/B scoring, and identical
wording and choices across labels, preventing question text from leaking
the answer. No video is duplicated in either evaluated subset. We do not
use temporal annotations, create multiple questions, or localize events;
multi-event identity is collapsed to the video-level positive label.
Paths absent on disk are skipped before deterministic interleaving.

\section{Qualitative Examples}
\label{app:qualfail}

\begin{figure}[h]
\centering
\includegraphics[width=\columnwidth]{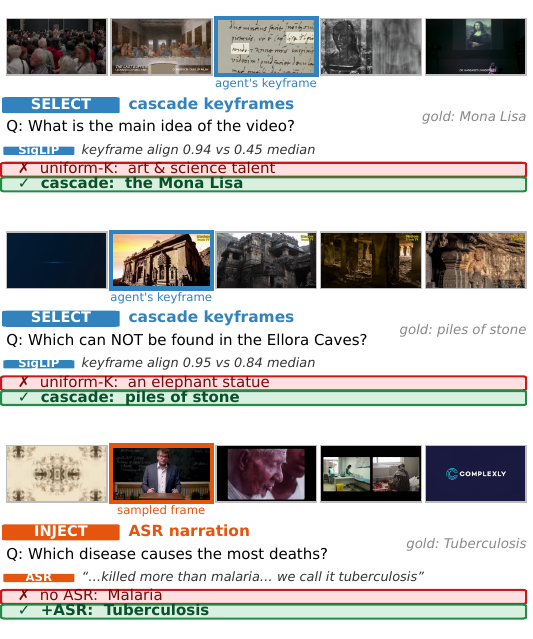}
\caption{\textbf{Qualitative wins on the SELECT and INJECT axes.}
Each panel shows a strip of the agent's sampled frames (top-scored
keyframe highlighted), the dispatched tool, and the answer (baseline
$\times$ $\to$ tool-routed \checkmark). The first two panels show SELECT
cases where cascade sampling surfaces an answer-bearing Video-MME frame;
the third shows an INJECT case where cached ASR supplies evidence absent
from the frames. All answers are from actual judged runs. These cases
illustrate mechanisms and are not representative of the average item.}
\label{fig:qualmain}
\end{figure}

\begin{figure}[h]
\centering
\includegraphics[width=\columnwidth]{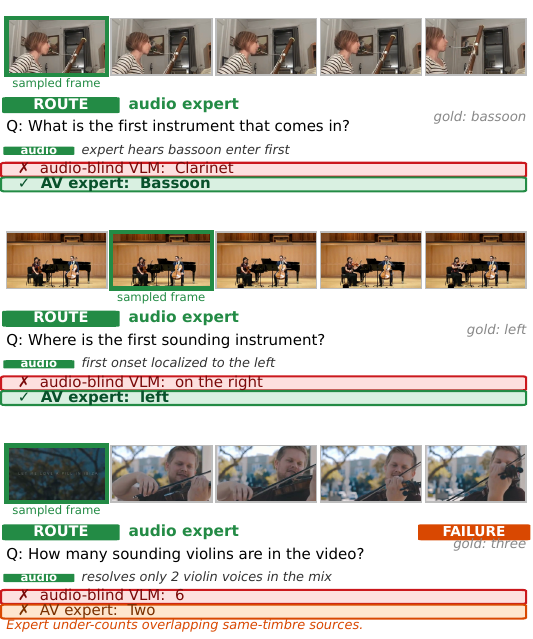}
\caption{\textbf{Answering-model successes and failure.}
The first two MUSIC-AVQA cases show the audio-visual expert correcting
answers from the audio-blind VLM; they illustrate the swap and do not
show that audio was needed (the silenced-soundtrack control is in the
extended F3(b) analysis below). In the third, the expert under-counts
overlapping same-timbre sources and remains wrong. All answers are from
actual scored runs; snippets are the cached read-outs.}
\label{fig:qualbottom}
\end{figure}

\section{Computational Cost Breakdown}
\label{app:cost}

\subsection*{Per-stage runtime in the audited runs (NVIDIA GB10, Qwen2.5-VL-7B)}
\begin{table}[t]
\centering
\caption{\textbf{Measured per-item runtime} (mean over the $n{=}900$
cascade$+$ASR run on one NVIDIA GB10, from the per-item timing records;
ms). EARS and uniform proposals, and SigLIP embeddings, were already
cached; model loading and the one-time preprocessing listed below are
excluded.}
\label{tab:coststage}
\small
\resizebox{\columnwidth}{!}{%
\begin{tabular}{lrl}
\toprule
Stage & ms/item & Notes \\
\midrule
Plan (rule-based)        & 0.1   & keyword match \\
Propose (EARS, uniform)  & 9     & cached proposals \\
Decode candidate frames  & 3{,}058 & $132$ frames, CPU \\
Filter (motion diff.)    & 15    & \\
Align (SigLIP)           & 18    & cached embeddings$^\ast$ \\
Select (facility loc.)   & 91    & \\
\midrule
Selection pipeline total & 3{,}220 & median $3{,}135$ \\
VLM call (8 frames$+$text) & 9{,}580 & median $10{,}750$ \\
\midrule
Wall clock per item      & 12{,}800 & uniform $10{,}100$; cascade $13{,}300$ \\
\bottomrule
\end{tabular}}

\smallskip\footnotesize $^\ast$In the selection-only cascade run, which
filled some of the cache, SigLIP alignment averaged $613$\,ms.
\end{table}

\paragraph{One-time preprocessing, excluded from the main-paper numbers.}
Whisper tiny.en transcription took $94$\,s per Video-MME long video on
average ($78$\,s median; $228$ videos, about $6$ GPU-hours as logged),
which is $\approx\!31$\,s per question at three questions per video. A
cold EARS proposal took $33$--$35$\,s on a $30$--$53$\,min video
(measured while other jobs shared the machine); its peaks fall within
the first $137$--$168$\,s. EasyOCR runs on the selected frames at
inference time and is not free: cascade$+$OCR took $21.6$\,s/item against
$13.3$ for the cascade (the per-item selection and VLM timers sum to only
$12.6$\,s, so the extra $\approx\!9$\,s is OCR extraction, including
filling its cache). Counting transcription, always-inject ASR costs about
$12.8+31\approx44$\,s per question, roughly four times uniform
sampling ($10.1$\,s), and the selection pipeline itself ($3.2$\,s,
dominated by decoding) adds about a third to the VLM call.

\paragraph{End-to-end timing protocol.}
The main paper reports mean wall-clock seconds per QA item,
\texttt{elapsed\_sec}/$n$, over each $n{=}900$ run on the same GB10.
Timing starts after model initialization and includes selection, cached
evidence lookup, image loading, the VLM call, and loop I/O. For every
run except cascade$+$OCR it equals the sum of the per-item selection and
VLM timers to within $0.05$\,s. The runs were separate and not isolated
from other jobs on the machine, so these are not isolated component
timings; run-level variance was not recorded.

\section{Method Details}
\label{app:method}
The main paper specifies proposal, filtering, alignment, selection and
reasoner routing. This section gives the configuration the audited runs
executed, the tool registry, the plan schema, the planner prompt, the
injection template and the selector ports.

\subsection*{Candidate pools: what the audited runs executed}
\label{app:pools}
A video is \emph{long} if it exceeds $120$\,s. The rule-based plan sets
$K_\text{proposal}{=}200$ (split evenly over the planned proposers),
$K_\text{keyframe}{=}32$, $K_\text{final}{=}K$, $\tau{=}2$\,s and a
uniform-anchor fraction of $0.5$. Table~\ref{tab:profiles} lists the two
planner profiles of the released code. The \emph{audit} profile is what
the published Video-MME long, XD-Violence and UCF-Crime cascade runs
executed.
\begin{table}[h]
\centering
\caption{\textbf{Planner profiles.} ``Planned'' is what the plan listed;
``pool'' is the number of candidate frames per long video the router
actually received (median over items; $64$ evenly spaced frames and no
filter for clips of $120$\,s or less).}
\label{tab:profiles}
\small
\resizebox{\columnwidth}{!}{%
\begin{tabular}{lcc}
\toprule
 & audit (published runs) & stacfp (released default) \\
\midrule
Planned proposers & shot, EARS, uniform & STACFP, EARS, uniform \\
Executed & EARS, uniform & STACFP, EARS, uniform \\
Pool (long video) & $132$ ($2{\times}66$) & $197$ ($3{\times}66$) \\
Filter & motion-difference & redundancy \\
Alignment & SigLIP (+ keyword aligners) & same \\
Select & $\tfrac{K}{2}$ facility-location + $\tfrac{K}{2}$ uniform & same \\
\bottomrule
\end{tabular}}
\end{table}
The plans of the audited runs listed a PySceneDetect shot-boundary
proposer, but PySceneDetect was not installed, so the proposer raised and
the router skipped it without a message (the shot-boundary cache files
first appear on the day it was installed; $132$ proposals is $2\times66$,
whereas three active proposers give $198$). We found this by re-running
the pipeline for release; the router now warns when a planned tool
fails, and the \texttt{audit} profile omits the proposer so that it
reproduces the published runs whether or not PySceneDetect is
installed (identical selected frames on $24$ of $24$ re-checked items; a
full re-run of $16$ items through the planner pipeline reproduced every
inject decision, injected-text length and prediction, and $12$ uniform
items reproduced every prediction). Two further
facts bound what the cascade can retrieve. (i)~The EARS proposer's motion
signal is computed from the first $4{,}096$ frames of the video (stride
$15$), so its event peaks lie within the first $\approx\!164$\,s of a
$25$\,fps video while its audio signal spans the whole clip. (ii)~The
anchors are $\mathrm{round}(K/2)$ evenly spaced frames chosen without
looking at any alignment score (Python rounding: $K{=}1,2,4,6,8$ give
$0,1,2,3,4$ anchors), so at $K{=}8$ the cascade decides only four frames.
The Qwen2-VL $n{=}900$ cascade$+$ASR run in our first cross-backbone
table was executed after PySceneDetect had been installed ($198$
proposals) while its cascade and uniform arms were not, so
Sec.~\ref{app:xvlm} does not interpret the ASR contrast there.

\paragraph{Uniform baselines.} \texttt{uniform} takes endpoint-inclusive
evenly spaced indices $\mathrm{round}(i\,(N{-}1)/(K{-}1))$, so it spends
two of eight frames on the first and last frame of the video and
the single frame at $K{=}1$ is frame $0$. \texttt{uniform\_center} takes
the midpoint of each of $K$ equal segments. At $K{=}8$ and $n{=}900$ it
scores $46.2\%$, identical to \texttt{uniform} ($0.0$\,pt, $95\%$ CI
$[-2.8,+2.9]$, $b/c{=}83/83$; $90\%$ CI $[-2.3,+2.3]$, equivalent within
$\pm3$\,pt), and against it the cascade is $+1.2$\,pt ($[-1.9,+4.3]$,
$p{=}0.49$) and the AKS port $+1.9$\,pt ($[-1.3,+5.1]$, $p{=}0.28$), both
equivalent within $\pm5$\,pt, so F1 does not depend on the endpoint-inclusive
spacing. At $K{=}1$ the two differ (Sec.~\ref{app:budget900}).

\subsection*{Tool registry}
\begin{table}[h]
\centering
\caption{\textbf{Registered tools} by pipeline stage, as exposed to the
planners.}
\label{tab:registry}
\small
\resizebox{\columnwidth}{!}{%
\begin{tabular}{lll}
\toprule
Tool & Stage & Behaviour \\
\midrule
\texttt{uniform\_stride} & Propose & evenly spaced frames, endpoints included \\
\texttt{ears\_event}     & Propose & audio-energy $+$ optical-flow event attention; top peaks \\
\texttt{shot\_boundary}  & Propose & PySceneDetect ContentDetector (threshold $30$); inactive in audited runs \\
\texttt{stacfp}          & Propose & colour-histogram clustering, cut-density adaptive \\
\texttt{motion\_diff}    & Filter & keep the $32$ frames with the largest grayscale change from their predecessor \\
\texttt{redundancy}      & Filter & drop near-duplicate frames by histogram similarity \\
\texttt{semantic\_siglip}& Align & SigLIP-base-patch16-224, max cosine over prompts \\
\texttt{object\_siglip}  & Align & SigLIP object/attribute prompts \\
\texttt{ocr\_text}       & Align & EasyOCR on candidate frames \\
\texttt{asr\_whisper}    & Align & Whisper tiny.en transcript overlap \\
\texttt{clap\_alm}, \texttt{audio\_energy} & Align & audio-language / loudness (not used in audited runs) \\
\bottomrule
\end{tabular}}
\end{table}

\subsection*{Plan schema and rule-based planner}
Both planners emit one JSON object with the keys
\texttt{proposers}, \texttt{filters}, \texttt{aligners} (lists of
registry names), \texttt{weights} (a map over the aligners, renormalized
to sum to one), \texttt{evidence\_prompts} (text prompts for the SigLIP
aligner), \texttt{K\_proposal}, \texttt{K\_keyframe}, \texttt{K\_final},
\texttt{tau\_sec}, \texttt{uniform\_anchor\_frac},
\texttt{inject\_text} ($\subseteq\{\texttt{asr},\texttt{ocr}\}$),
\texttt{expected\_expert} and \texttt{reasoning}. The rule-based planner
lowercases the question and, by substring match, adds
\texttt{ocr\_text} (weight $0.5$) for \emph{text, sign, subtitle,
license plate, letter, word, written}; \texttt{asr\_whisper} ($0.4$) for
\emph{say, said, speak, spoken, talk, dialog(ue), conversation, shout,
yell}; and \texttt{object\_siglip} ($0.4$) for \emph{color, wearing,
holding, on the table, in the room, near the, next to, behind, in front
of}. \texttt{semantic\_siglip} always gets $\max(0.3,\,1-\sum\text{others})$.
The evidence prompt defaults to the first $80$ characters of the
question. The rule-based planner never sets \texttt{inject\_text}.

\subsection*{LLM planner prompt}
The planner is Qwen2.5-3B-Instruct with greedy decoding and at most
$384$ new tokens; malformed JSON falls back to the rule-based plan. In
\texttt{inject\_only\_mode} only the \texttt{inject\_text} field of the
reply is used. The prompt of the audited runs is, in order: the
instruction ``\emph{You are an agentic video understanding planner. Pick
the MINIMAL set of tools that answers the question; every extra tool
costs compute. The VLM will see the selected frames as pixels. If you set
inject\_text=["asr"] the cached speech transcript near the selected
frames will ALSO be inserted into the VLM prompt as text (highly
effective when the answer is spoken). Use inject\_text=["ocr"] for
on-screen text questions.}''; the memory block (Sec.~\ref{app:procmem})
under ``\emph{PROCEDURAL MEMORY (priors learned from past execution logs;
treat as strong defaults, not absolute rules)}''; the question, the
dataset id and the duration tagged LONG/SHORT; the list of available
tools with relative cost hints; six guidelines (always include
\texttt{uniform\_stride} and \texttt{semantic\_siglip}; add
\texttt{object\_siglip}, \texttt{asr\_whisper} or \texttt{ocr\_text} only
for object, speech or on-screen-text questions; $1$--$3$ aligners;
weights sum to one); three few-shot plans (a colour question with
\texttt{inject\_text=[]}, a narration question with \texttt{["asr"]}, a
sign question with \texttt{["ocr"]}); and the request to reply with one
JSON object. The exact text is \texttt{LLMCascadePlanner.\_build\_prompt\_paper}
in the released code. The prompt in the code of the first public
snapshot additionally described STACFP, CLAP and audio-energy tools; the
released planner defaults to the audited prompt (\texttt{prompt\_version="paper"}).

\subsection*{Injection prompt template}
The \emph{inject} stage appends a text block to the VLM prompt \emph{after}
the question and options. The multiple-choice prompt is
\begin{quote}\small
Question: \dots\\
Options:\\
A.\ \dots\ \ B.\ \dots\ \ \dots\\
Answer with the single letter (A/B/C/D) of the best option, and nothing
else.
\end{quote}
followed, when text is injected, by
\begin{quote}\small
Additional evidence extracted from the video (treat as context, not
ground truth):\\
Transcript near the selected frames:\\
- [mm:ss] \textit{passage ($\le240$ chars)}\ \dots\\
Overall transcript (truncated): \textit{summary ($\le400$ chars)}\\
\textit{[On-screen text detected in the selected frames:}\\
\textit{- [mm:ss] text ($\le160$ chars per frame)]}
\end{quote}
Passages are the cached Whisper chunks overlapping $\pm4$\,s of a selected
frame. Over the $n{=}900$ runs ASR injection adds on average $1{,}344$
characters (median $1{,}320$; $95$th percentile $1{,}724$; maximum
$5{,}053$) and OCR $274$. The block is built from cached
transcripts: a video without one receives no text, and the harness
reports how many items were injected (\texttt{qa\_summary.json},
\texttt{injection}). Injection is \emph{ablatable} and matched in visual
budget and decoding configuration; the total context is deliberately not
matched, because the text is the intervention under test. Cached evidence
lookup is negligible relative to the VLM forward pass; the one-time
extraction cost is excluded as described in Sec.~\ref{app:cost}.

\subsection*{Selector ports}
\label{app:ports}
All three ports consume the same front end: SigLIP-base image and text
embeddings of a pool of $256$ frames whose stride makes them span the
entire video (an earlier pool covered only the first $\approx\!42$\,s of
each video; the published port rows are from the re-run after that fix),
and relevance is the maximum cosine similarity to the question prompt.
\textbf{AKS}: relevance is min--max normalized, the official bisection
runs with $t_1{=}0.8$, $t_2{=}-100$ and depth $5$, and each terminal
segment of depth $d$ contributes its top $\max(1,\lfloor K/2^d\rfloor)$
frames; the result is trimmed to $K$ by score or padded with evenly
spaced frames. \textbf{KeyScore}: $\alpha{=}0.85$ (semantic),
$\beta{=}0.05$ (similarity to the mean-pooled video embedding) and
$\gamma{=}0.10$ (drop impact, in closed form), each min--max normalized,
top $K$ by score. \textbf{VideoTree}: $K$-means ($K$ clusters, $4$
initializations, seed $0$) over the image embeddings, clusters ranked by
peak relevance, the most relevant frame of each of the top $K$ clusters;
its iterative LLM-guided expansion and captions are not ported.
\texttt{scripts/port\_fidelity.py} compares the rules with their
references on identical score arrays. On $100$ real SigLIP score tracks
from Video-MME long (a $256$-frame pool each) the KeyScore port matches
the reference implementation's scores to $3.5\times10^{-5}$ and selects
the identical top-$8$ set on all $100$; its weights are the reference
\texttt{FrameScorer} defaults, and the reference API's current defaults
($0.70/0.10/0.20$) change $4\%$ of the top-$8$ set. For AKS the official
routine, run at its default depth $5$ with $K{=}8$, returns \emph{no
frame at all} on $100$ of $100$ tracks: no segment reaches the
$t_1{=}0.8$ peakiness threshold on min--max-normalized scores, so every
segment splits to depth $5$, where $\lfloor 8/2^5\rfloor{=}0$ frames
remain. The official routine is designed for budgets of $32$--$64$
frames. The port therefore differs by construction: it is a temporally
binned top-$K$ (the best frame in each of $32$ depth-$5$ bins, trimmed to
the best $K{=}8$ by score), which overlaps the naive top-$8$ by a Jaccard
index of $0.66$ and the official routine at depth $3$ (the deepest
setting that still emits $8$ frames) by $0.47$. VideoTree has no released
selection routine to compare against. The AKS and VideoTree rows
therefore bound a \emph{selection rule adapted to $K{=}8$}, not the
published systems.

\section{Extended Experiments}
\label{app:expt}
This section holds the detailed results condensed in
the main paper's experiments: the subset-scale selector ablation, the
selection$\times$injection factorial, judge validation, the full
statistical-robustness table, cross-model context, and the extended
expert-routing analysis.

\subsection*{Subset-scale selector ablation ($n{=}200$)}
Table~\ref{tab:eventsparse} is the matched visual-budget selector
comparison at our $n{=}200$ development scale.
Single-cue content-aware selectors do not beat uniform sampling, and
the forced multi-cue selector does not improve over the anchored
cascade. This subset is underpowered; the full-bucket comparison and
formal equivalence tests are reported in the main paper and below.

\begin{table}[h]
\centering
\caption{Subset-scale selector ablation. Matched-budget selectors on
Video-MME long ($n{=}200$, $K{=}8$, Qwen2.5-VL-7B). No selector
separates from uniform at this scale; the forced multi-cue planner costs
$5\times$ more without a significant gain.}
\label{tab:eventsparse}
\small
\resizebox{\columnwidth}{!}{%
\begin{tabular}{lcccc}
\toprule
Selector ($K{=}8$) & $n$ & Acc. (\%) & Wall s/item & $p$ vs.\ ours \\
\midrule
Random-$K$                          & 200 & 48.5 & 10.0 & 0.154 \\
Uniform-$K$                         & 200 & 50.0 & 10.0 & 0.324 \\
Motion (motion-diff only)           & 200 & 46.5 & 20.1 & 0.049 \\
CLIP (SigLIP only)                  & 200 & 49.5 & 20.9 & 0.312 \\
Forced multi-cue (Obj$+$ASR$+$SigLIP) & 200 & 50.5 & 76.9 & 0.327 \\
\textbf{Cascade (anchored relevance)} & 200 & \textbf{53.5} & 14.3 & --- \\
\bottomrule
\end{tabular}}
\end{table}

\subsection*{Selection $\times$ injection factorial ($n{=}200$)}
\begin{figure}[h]
\centering
\includegraphics[width=\columnwidth]{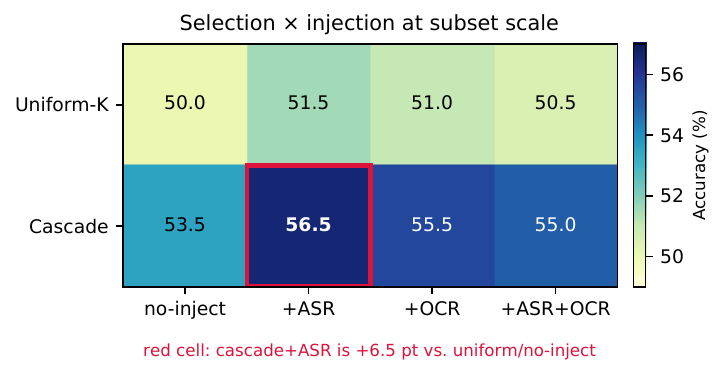}
\caption{Apparent selection $\times$ injection interaction at
$n{=}200$. Accuracy (\%) on Video-MME long ($K{=}8$,
Qwen2.5-VL-7B). ASR and OCR are injected into the VLM prompt. OCR adds
$+2.0$\,pt over cascade-no-inject but is redundant once ASR is present;
stacking both scores below either alone.}
\label{fig:f2heatmap}
\end{figure}

Figure~\ref{fig:f2heatmap} reports the controlled $2{\times}4$ grid on
Video-MME long. At this scale, always-inject ASR adds $+1.5$\,pt over
uniform frames and $+3.0$\,pt over cascade frames. The combined
cascade$+$ASR cell is $+6.5$\,pt above uniform-no-inject, but this
attenuates to $+1.7$\,pt on the full $n{=}900$ bucket. The grid is
therefore illustrative rather than evidence of a stable interaction.

\subsection*{Anomaly-domain ASR, and where the text sits in the prompt}
\label{app:anomasr}
\paragraph{What the published anomaly arms did.} The $n{=}200$
``cascade$+$ASR'' arms on XD-Violence and UCF-Crime matched the plain
cascade on every item: the evidence-text length recorded for all $400$
questions is zero, there are $0$ discordant pairs against the cascade
($91.0{\to}91.0$ and $85.0{\to}85.0$), and the per-question planner
chose \texttt{inject\_text=[]} for all $400$ questions. No Whisper
transcript had been computed for those clips, and the injection code
silently injects nothing for a video without one; it now warns and
reports how many items received text. Every number that attributes an
anomaly-domain gain to ``$+$ASR'' in earlier versions of this work is
therefore a selection effect.

\paragraph{Re-run with transcripts.} We transcribed all $400$ clips with
Whisper tiny.en (CPU, five shards) and re-ran the arm with the published
configuration (audit profile, $K{=}8$, $\pm4$\,s windows, same prompt).
Every question received text ($1.6$k characters on average). Table~\ref{tab:placement}
reports it next to the same arm with the identical block placed
\emph{before} the question (\texttt{--evidence\_position before}); frames,
transcripts, backbone and decoding are unchanged between the two.
Appended after the options and the answer instruction, ASR lowers
XD-Violence from $91.0$ to $81.0\%$ ($b/c{=}2/22$); the loss is entirely
on the violent class ($62$ of $100$ Fighting clips correct against $83$
without text, $100$ of $100$ Normal clips against $99$): the model
reads the speech as evidence of an ordinary scene. Placed before the
question, the same text leaves XD at the cascade's accuracy. On
UCF-Crime both placements give $72.0\%$ against $85.0\%$ without text
($b/c{=}1/27$, identical predictions on $190$ of $200$ clips); the
violent clips are the ones that flip to ``no unusual activity''
($55$ of $100$ answered ``No'' against $28$ without text).

\paragraph{Why UCF-Crime differs.} The UCF-Crime clips are silent
surveillance footage, and Whisper tiny.en fills silence with its
characteristic hallucination: $160$ of the $200$ transcripts contain
``thank you for watching'' at least three times and $190$ are dominated by
repetition (fewer than $20\%$ distinct words), against $9$ and $126$ of
$200$ for XD-Violence. The injected ``evidence'' there is mostly artefact,
and it pushes answers toward ``no unusual activity'' in either
placement. We did not filter hallucinated transcripts (for example with
voice-activity detection), so these arms measure the injection protocol as
audited, not the best attainable ASR evidence.

\begin{table}[h]
\centering
\caption{\textbf{Evidence placement.} Qwen2.5-VL-7B, $K{=}8$, audit-profile
cascade; accuracy (\%), $\Delta$ is the paired difference to the cascade
without text with its $95\%$ bootstrap interval and exact McNemar $p$.
\emph{After} is the audited protocol (block appended after the question,
options and answer instruction); \emph{before} puts the identical block ahead
of the question. $^\dagger$The published anomaly arms injected nothing.}
\label{tab:placement}
\small
\setlength{\tabcolsep}{3pt}
\resizebox{\columnwidth}{!}{%
\begin{tabular}{llcccc}
\toprule
Domain & Arm & $n$ & Acc. & $\Delta$ vs.\ cascade & $b/c$, $p$ \\
\midrule
XD F/N & Cascade, no text & 200 & 91.0 & --- & --- \\
       & $+$ASR, as published$^\dagger$ & 200 & 91.0 & $0.0$ & $0/0$, $1$ \\
       & $+$ASR, after the question & 200 & 81.0 & $-10.0$ $[-14.5,-5.5]$ & $2/22$, $<10^{-4}$ \\
       & $+$ASR, before the question & 200 & 91.0 & $0.0$ $[-3.0,+3.0]$ & $5/5$, $1$ \\
\midrule
UCF-Crime & Cascade, no text & 200 & 85.0 & --- & --- \\
       & $+$ASR, as published$^\dagger$ & 200 & 85.0 & $0.0$ & $0/0$, $1$ \\
       & $+$ASR, after the question & 200 & 72.0 & $-13.0$ $[-18.0,-8.5]$ & $1/27$, $<10^{-5}$ \\
       & $+$ASR, before the question & 200 & 72.0 & $-13.0$ $[-18.0,-8.5]$ & $1/27$, $<10^{-5}$ \\
\midrule
Video-MME long & Cascade, no text & 900 & 47.4 & --- & --- \\
       & $+$ASR, after the question & 900 & 47.9 & $+0.4$ $[-1.6,+2.4]$ & $45/41$, $0.75$ \\
       & $+$ASR, before the question & 900 & 50.2 & $+2.8$ $[+0.1,+5.6]$ & $91/66$, $0.055$ \\
\bottomrule
\end{tabular}}
\end{table}

\paragraph{Consequences.} The planner arms on the anomaly clips never
injected, so they say nothing about injection there. The audited
Video-MME injection arms all use the appended placement; the table shows
what the alternative placement does on the same selected frames, and the
main paper treats the sign of an injection effect as protocol-dependent.
On Video-MME the before-question arm injected the same text into the same
$865$ of $900$ questions as the appended arm (identical evidence length on
all $900$), so the two differ in placement alone. Placing the text first
is $+2.3$\,pt over appending ($[-0.2,+4.9]$, $p{=}0.087$) and $+4.0$\,pt
over uniform sampling ($[+0.7,+7.3]$; video-clustered $[+0.9,+7.2]$;
exact McNemar $p{=}0.024$, $b/c{=}139/103$). Its gain over the no-text
cascade is positive in each part of the bucket ($+3.5$, $+4.8$ and
$+1.3$\,pt on items $1$--$200$, $201$--$450$ and $451$--$900$; $p{=}0.30$,
$0.13$, $0.55$). Uniform sampling with the text first, run afterwards,
scores $50.7\%$: $+4.4$\,pt over plain uniform ($[+1.8,+7.2]$, $p{=}0.002$,
$b/c{=}100/60$) and indistinguishable from the cascade with the text first
($p{=}0.83$, $b/c{=}91/95$), so the gain comes from the text and not from
the cascade's frames. A four-position control with uniform frames
(Table~\ref{tab:fourpos}) shows the gain is specific to putting the text
\emph{before the question}: between the question and the options it is
$+1.0$\,pt, between the options and the answer instruction $+1.9$\,pt and
after the instruction (audited) $+1.6$\,pt, none significant, while before
the question it is $+4.4$\,pt. The two other positions do not differ from
the audited one ($p{\ge}0.75$), whereas before vs.\ between question and options
is $+3.4$\,pt ($[+1.0,+6.0]$, $p{=}0.008$). All four prompts carry the same
$900$ evidence blocks (identical evidence length on every item), and all but
one of the $3{,}600$ answers across the four arms start with an option letter,
so the differences are not format failures. On Qwen2-VL-7B (same $900$ items, uniform
sampling, $45.4\%$) the text first gives $52.7\%$ ($+7.2$\,pt,
$[+4.4,+9.9]$, $p{<}10^{-6}$, $b/c{=}111/46$) and the appended text
$47.2\%$ ($+1.8$\,pt, $[-0.1,+3.8]$, $p{=}0.093$); first vs.\ appended is
$+5.4$\,pt ($[+2.7,+8.3]$, $p{=}0.0002$). Placement was varied after the
anomaly results and these contrasts are unadjusted (the Bonferroni level
for four is $0.0125$, which uniform text-first vs.\ no text, $p{=}0.002$, and
before vs.\ between question and options pass; before vs.\ audited,
$p{=}0.038$, does not).

\begin{table}[h]
\centering
\caption{\textbf{Where the text goes: four prompt positions} (Qwen2.5-VL-7B,
uniform frames, $K{=}8$, $n{=}900$, same ASR block). $\Delta$ is the paired
difference to uniform without text, with exact McNemar $p$.}
\label{tab:fourpos}
\small
\setlength{\tabcolsep}{4pt}
\begin{tabular}{lccc}
\toprule
Position of the ASR block & Acc. & $\Delta$ & $p$ \\
\midrule
No text                              & 46.2 & --- & --- \\
After the answer instruction (audited) & 47.8 & $+1.6$ & 0.15 \\
Between options and instruction      & 48.1 & $+1.9$ & 0.082 \\
Between question and options         & 47.2 & $+1.0$ & 0.57 \\
Before the question                  & \textbf{50.7} & $+4.4$ & 0.002 \\
\bottomrule
\end{tabular}
\end{table}
The pooled analysis of Sec.~\ref{app:expt} (stratified McNemar over the
three domains) is recomputed with the transcribed arms by
\texttt{scripts/pooled\_sensitivity.py}: with the transcribed arms the
pooled cascade$+$ASR contrast over the three $n{=}200$ subsets is
$-3.0$\,pt ($95\%$ CI $[-6.2,0.0]$; stratified McNemar $p{=}0.063$; per
domain $+6.5$, $-6.0$ and $-9.5$\,pt), and weighting the three domains
equally gives $-4.6$\,pt ($[-7.1,-2.1]$, sign-flip $p{=}0.0003$): once the
anomaly arms actually inject, the pooled gain disappears. The pooled
within-cascade ASR effect is $-6.7$\,pt ($[-9.3,-4.2]$) on that pool and
$-3.2$\,pt ($[-5.0,-1.5]$) with all $900$ Video-MME items.

\subsection*{Budget sweep and the $K{=}1$ result}
\label{app:budget900}
\begin{table}[h]
\centering
\caption{\textbf{Budget sweep at $n{=}200$.} Cascade vs.\ uniform on
Video-MME long (no inject, Qwen2.5-VL-7B); $p$ is two-sided exact
McNemar, unadjusted across the five budgets. Accuracy need not be
monotone in $K$: decoding is greedy, so the $K{=}6{\to}8$ dip is
deterministic and consistent with context dilution, although the
experiment does not isolate its cause.}
\label{tab:ksweep}
\small
\setlength{\tabcolsep}{4pt}
\begin{tabular}{lccccc}
\toprule
$K$ & 1 & 2 & 4 & 6 & 8 \\
\midrule
Uniform-$K$ & 38.0 & 43.0 & 49.0 & \textbf{52.5} & 50.0 \\
Cascade     & \textbf{48.0} & \textbf{47.0} & \textbf{50.0} & 51.0 & \textbf{53.5} \\
$\Delta$    & $+10.0$ & $+4.0$ & $+1.0$ & $-1.5$ & $+3.5$ \\
$p$         & \textbf{0.009} & 0.302 & 0.875 & 0.728 & 0.324 \\
\bottomrule
\end{tabular}
\end{table}
Two properties of the baselines matter for the $K{=}1$ cell. The
cascade reserves $\mathrm{round}(K/2)$ uniform anchors, which is $0$ at
$K{=}1$ (Sec.~\ref{app:pools}), so at $K{=}1$ the cascade is a pure
selector while at every larger budget half of its frames are uniform.
And the endpoint-inclusive uniform baseline places its single frame at
frame of the video, which is often a title or black frame. A full-scale
$K{=}1$ run ($n{=}900$, Qwen2.5-VL-7B) confirms both points: uniform $37.0\%$,
cascade $39.7\%$ ($+2.7$\,pt, $[-0.7,+6.0]$, $p{=}0.13$, $b/c{=}126/102$),
and the middle-frame baseline (\texttt{uniform\_center}) $42.1\%$, which is
$+5.1$\,pt over first-frame uniform ($[+1.9,+8.4]$, $p{=}0.003$) and
$2.4$\,pt above the cascade ($[-5.7,+0.8]$, $p{=}0.16$). The $+10.0$\,pt
of the $n{=}200$ prefix is thus a subset estimate against a weak baseline
(the prefix gain is $3.7\times$ the full-scale one).

\subsection*{Statistical robustness (F1 null + F3 effects)}
We are explicit that a non-significant $p$-value is not by itself
evidence of equivalence. The ``Pow.'' column of Table~\ref{tab:stats} is
post-hoc power computed at the \emph{observed} effect, which is a
monotone transform of the $p$-value and is reported for completeness
only; the evidence we rely on is the discordant-pair count, the paired
CI, the MDE and the TOST verdict. At $n{=}200$ the forced-multi-cue-vs-uniform
contrast has $95\%$ CI $[-5.5,+6.5]$\,pt, so
the subset is underpowered to certify the null (its MDE at $80\%$ power
is $8.3$\,pt; see Sec.~\ref{app:mde} for the derivation and for the
per-contrast values). The formal equivalence evidence therefore comes from the full
$n{=}900$ bucket, where selection-only cascade sits at $+1.2$\,pt over
uniform ($95\%$ CI $[-1.7,+4.1]$) and the per-question planner is
statistically equivalent to always-inject ASR ($+0.6$\,pt, $90\%$ CI
$[-0.9,+2.0]$ within a $\pm3$\,pt margin).

\paragraph{Equivalence-margin sensitivity.} The $\pm5$\,pt margin used
in the main paper is set to the top of the range of selection-axis gains
this literature claims ($1$--$5$\,pt), so passing it means the full
bucket rules out a literature-scale improvement. It is not a margin the
data chose: at $\pm3$\,pt \emph{no} full-bucket selection contrast
passes (cascade $90\%$ CI $[-1.2,+3.7]$; AKS $[-0.8,+4.6]$; KeyScore
$[-1.7,+3.4]$; VideoTree $[-1.1,+4.1]$; cascade$+$OCR $[-1.6,+3.4]$).
We therefore distinguish three verdicts throughout: failure to reject a
difference ($p{>}0.05$), demonstrated equivalence at a stated margin
($\pm5$\,pt here), and insufficient precision for a tighter margin
($\pm3$\,pt here).

\paragraph{Where the $1$--$5$\,pt range comes from.}
Table~\ref{tab:litscale} transcribes selection-axis contrasts from the
published tables of the two selectors we port, together with the size
of the evaluation set each number is measured on. Headline
selector-vs.-uniform gains on Video-MME are $0.9$--$2.3$\,pt, and the
component-level contrasts these papers report on few-hundred-item
evaluation sets span $0.2$--$5.0$\,pt. The single value above
$5$\,pt compares VideoTree against LLoVi and also changes the reasoner
and the caption budget, so it is not a selection-only number. This is
the evidence behind the $1$--$5$\,pt effect range we quote and behind
the $\pm5$\,pt equivalence margin, and it is why the $n{=}200$ MDE of
$8.3$--$10.4$\,pt (Sec.~\ref{app:mde}) is the central problem: the
effects this literature reports are smaller than the evaluation sizes
it uses can resolve.

\begin{table}[t]
\centering
\caption{\textbf{Reported selection-axis contrasts and the evaluation
sizes they are measured on.} Values are transcribed from the cited
papers' own tables; $n$ is the number of scored questions in the
evaluation set that number comes from, and $K$ is the frame budget
given to the answering model. Each gain is against that paper's own
baseline and is not comparable across rows. The VideoTree rows contrast
the full pipeline, or one ablated component of it, against that paper's
baseline, so they are not pure selection-rule swaps.}
\label{tab:litscale}
\small
\resizebox{\columnwidth}{!}{%
\begin{tabular}{llrlr}
\toprule
Source & Contrast & $n$ & $K$ & Gain \\
\midrule
AKS~\cite{tang2025aks} T1  & vs.\ uniform, Qwen2-VL-7B      & 2700 & 32 & $+2.3$ \\
AKS~\cite{tang2025aks} T1  & vs.\ uniform, LLaVA-OV-7B      & 2700 & 32 & $+1.9$ \\
AKS~\cite{tang2025aks} T2  & ADA vs.\ UNI, LLaVA-Video-7B   & 2700 & 64 & $+0.9$ \\
VideoTree~\cite{wang2024videotree} T2  & vs.\ LLoVi, Video-MME long     & 900 & adapt. & $+5.4$ \\
VideoTree~\cite{wang2024videotree} T11 & vs.\ uniform, EgoSchema sub.   & 500 & adapt. & $+5.0$ \\
VideoTree~\cite{wang2024videotree} T5  & depth-expansion component      & 500 & adapt. & $+1.8$ \\
VideoTree~\cite{wang2024videotree} T10 & 8B vs.\ 88M vision encoder     & 500 & adapt. & $+0.2$ \\
\midrule
\textbf{This paper} & best port vs.\ uniform, V-MME long & 900 & 8 & $+1.9$ \\
\bottomrule
\end{tabular}}
\end{table}

\paragraph{Video-clustered re-analysis.} Video-MME long asks three
questions per video, so item-level resampling could understate
uncertainty. We recompute every full-bucket interval with a
video-clustered bootstrap that resamples the $300$ videos with
replacement rather than the $900$ items, and repeat the F2 subset
experiment by drawing $67$ of $300$ \emph{videos} (the same
$\approx\!200$ questions) instead of $200$ items. The effect is
negligible: the clustered bootstrap standard deviation is
$0.996$--$1.02\times$ the item-level one, every $95\%$ CI moves by
$\le\!0.3$\,pt, and no TOST verdict changes (cascade vs.\ uniform
clustered $95\%$ CI $[-1.8,+4.1]$; cascade$+$ASR $[-1.7,+4.9]$; planner
vs.\ always-ASR $[-1.2,+2.3]$). The clustered subset draws give a
$95\%$ range of $[-4.0,+6.5]$\,pt with a $35.7\%$ sign-flip rate for
cascade vs.\ uniform, against $[-4.0,+6.5]$ and $36.0\%$ item-level.
Within-video outcome correlation is thus not what drives F2. The
anomaly domains contribute one question per video, so no clustering
correction applies to them. Script:
\texttt{scripts/cluster\_robustness.py}.

\paragraph{The pooled stratified test.} The $n{=}600$ result in the main
paper is a Mantel--Haenszel-style stratified McNemar over three strata
(one per domain), weighted equally by item:
$Z=\bigl(\sum_k b_k-\sum_k c_k\bigr)/\sqrt{\sum_k(b_k{+}c_k)}$ with
$b_k,c_k$ the discordant counts of stratum $k$. For cascade$+$ASR vs.\
uniform-no-inject the strata are $b/c=28/15$ (Video-MME long),
$14/6$ (XD-Violence F/N) and $12/5$ (UCF-Crime), giving $b{=}54$, $c{=}26$,
$Z{=}3.13$ and $p{=}0.0017$ two-sided; with a continuity correction
$p{=}0.0025$. The paired deltas are $+6.5$, $+4.0$ and $+3.5$\,pt and no
single domain is significant on its own ($p{=}0.066$, $0.115$,
$0.144$)---the pooled test is what carries F3(a). The $95\%$ CI
$[+1.8,+7.7]$ is a stratified percentile bootstrap resampling within
domains.

\paragraph{Sensitivity of the pooled estimate to the development prefix.}
The published pool takes the first $200$ Video-MME items, the prefix on
which the cascade and the planner memory were developed, plus the two
$200$-item anomaly sets. Replacing the prefix by all $900$ items
($n{=}1{,}300$) gives $+2.31$\,pt ($95\%$ CI $[-0.1,+4.7]$;
$b/c=140/110$, $Z{=}1.90$, $p{=}0.058$); using only items outside the
prefix ($700$ Video-MME and $400$ anomaly items, $n{=}1{,}100$) gives
$+1.55$\,pt ($[-1.1,+4.1]$; $112/95$, $p{=}0.24$). The $700$ held-out
Video-MME items alone show $+0.29$\,pt ($86/84$, exact $p{=}0.94$). The
selection-only cascade follows the same pattern ($+3.67$\,pt on the
published pool, $+2.00$ on all data, $+1.73$ on held-out items). Weighting the three domains equally
($1/3$ each, within-domain bootstrap, sign-flip test) gives $+3.06$\,pt
($[+0.9,+5.3]$, $p{=}0.0076$) with all $900$ Video-MME items and
$+2.60$\,pt ($[+0.3,+5.0]$, $p{=}0.033$) without the prefix, because the
two anomaly domains carry the largest point estimates and receive $2/3$
of the weight. Within the cascade, ASR adds $+3.00$\,pt on the prefix
($10/4$, $p{=}0.18$) and $-0.29$\,pt on the other $700$ items ($35/37$,
$p{=}0.91$). Script: \texttt{scripts/pooled\_sensitivity.py}.

\begin{table}[h]
\centering
\caption{Paired $\Delta$ (pt), 95\% CI, McNemar $p$, power. Formal null
from the $n{=}900$ bucket.}
\label{tab:stats}
\small
\resizebox{\columnwidth}{!}{%
\begin{tabular}{lcccccc}
\toprule
Contrast ($K{=}8$) & $n$ & $\Delta$ & 95\% CI & $p$ & Pow. & Equiv.\ \\
\midrule
\multicolumn{7}{l}{\emph{F1 --- selection-only is null (equivalence wanted)}}\\
\;Motion vs uniform        & 200 & $-3.5$ & $[-9.5,+3.0]$ & 0.349 & 0.19 & -- \\
\;Clip vs uniform          & 200 & $-0.5$ & $[-8.0,+6.5]$ & 1.000 & 0.05 & -- \\
\;Forced-multicue vs unif. & 200 & $+0.5$ & $[-5.5,+6.5]$ & 1.000 & 0.05 & -- \\
\;Cascade vs uniform       & 200 & $+3.5$ & $[-2.5,+9.5]$ & 0.324 & 0.21 & n/a \\
\midrule
\multicolumn{7}{l}{\emph{F3(a) --- injection effect (effect wanted)}}\\
\;Uniform: $+$ASR          & 200 & $+1.5$ & $[-2.0,+5.5]$ & 0.607 & 0.12 & n/a \\
\;Cascade: $+$ASR          & 200 & $+3.0$ & $[-0.5,+6.5]$ & 0.180 & 0.36 & n/a \\
\;Headline casc$+$ASR/unif & 200 & $+6.5$ & $[+0.5,+13.0]$ & 0.066 & 0.51 & n/a \\
\;ASR$+$OCR vs ASR         & 200 & $-1.5$ & $[-4.0,+0.5]$ & 0.375 & 0.27 & \checkmark \\
\midrule
\multicolumn{7}{l}{\emph{Full bucket $n{=}900$ (formal null evidence)}}\\
\;Selection-only casc/unif & 900 & $+1.2$ & $[-1.7,+4.1]$ & 0.452 & 0.13 & $\pm5$ \\
\;Cascade$+$ASR vs uniform & 900 & $+1.7$ & $[-1.6,+4.8]$ & 0.337 & 0.18 & n/a \\
\;Planner vs uniform       & 900 & $+2.2$ & $[-0.7,+5.1]$ & 0.154 & 0.32 & n/a \\
\;Planner vs always-ASR    & 900 & $+0.6$ & $[-1.2,+2.2]$ & 0.615 & 0.10 & \checkmark \\
\bottomrule
\end{tabular}}
\end{table}

\subsection*{Minimum detectable effect: derivation and per-contrast values}
\label{app:mde}
McNemar's test conditions on the \emph{discordant} pairs---items that one
arm answers correctly and the other does not. Concordant items (both
correct, or both wrong) carry no information about which arm is better,
so the effective sample size of a paired comparison is not $n$ but
$b{+}c$. Writing $p_{\text{disc}}{=}(b{+}c)/n$, the variance of the
paired accuracy difference is approximately $p_{\text{disc}}/n$, and the
usual normal approximation to McNemar power gives a minimum detectable
effect
\begin{equation}
\mathrm{MDE} \;=\; \bigl(z_{1-\alpha/2} + z_{\beta}\bigr)
\sqrt{\frac{p_{\text{disc}}}{n}}\times 100 \quad\text{pt},
\label{eq:mde}
\end{equation}
with $z_{0.975}{=}1.960$ and $z_{0.80}{=}0.842$ at $\alpha{=}0.05$
(two-sided) and $80\%$ power. Worked example for the
forced-multi-cue-vs-uniform contrast quoted in the main text: $b{=}18$
and $c{=}17$, so $35$ of $200$ items are discordant
($p_{\text{disc}}{=}0.175$) and
$\mathrm{MDE}{=}2.802\sqrt{0.175/200}\times100{=}8.3$\,pt.

Because the MDE scales with $\sqrt{p_{\text{disc}}}$ it is a property of
a \emph{contrast}, not of $n$ alone: two arms that seldom disagree can be
separated at a smaller effect size than two arms that disagree often.
Table~\ref{tab:mde} therefore reports $b$, $c$ and the resulting MDE for
every contrast we run rather than a single number. The pattern that
matters for F1 and F2 is that all four selection-axis contrasts at
$n{=}200$ sit between $8.3$ and $10.4$\,pt, while every selection-axis
gain reported in this literature is $1$--$5$\,pt; moving to $n{=}900$
roughly halves the MDE, to $4.1$--$4.5$\,pt. The injection contrasts have
much lower discordant rates ($2.5$--$7.5\%$) and correspondingly smaller
MDEs, which is why a $+3.0$\,pt injection effect is more informative at
the same $n$ than a $+3.5$\,pt selection effect.

An MDE is not a significance threshold. It is the effect size at which a
study would reject the null $80\%$ of the time; smaller effects are not
undetectable, only more often missed than found. The consequence for
reading this literature is asymmetric and worth stating plainly: at
$n{=}200$ on the selection axis, a report of ``no significant
difference'' establishes very little. Significant subset estimates can
also be unstable and should be read with paired intervals and MDEs.

\begin{table}[h]
\centering
\caption{Discordant-pair counts and minimum detectable effect
(Eq.~\ref{eq:mde}; $80\%$ power, $\alpha{=}0.05$). $b$ counts items only
the second arm answers correctly, $c$ only the first; just $b{+}c$
carries information. Computed from cached per-item predictions.}
\label{tab:mde}
\small
\setlength{\tabcolsep}{4pt}
\begin{tabular}{lccccc}
\toprule
Contrast ($K{=}8$) & $n$ & $b$ & $c$ & $p_{\text{disc}}$ & MDE \\
\midrule
\multicolumn{6}{l}{\emph{Selection axis}}\\
\;Motion vs uniform        & 200 & 17 & 24 & 20.5\% & 9.0 \\
\;CLIP vs uniform          & 200 & 27 & 28 & 27.5\% & 10.4 \\
\;Forced-multicue vs unif. & 200 & 18 & 17 & 17.5\% & 8.3 \\
\;Cascade vs uniform       & 200 & 22 & 15 & 18.5\% & 8.5 \\
\;Cascade vs uniform       & 900 & 94 & 83 & 19.7\% & 4.1 \\
\midrule
\multicolumn{6}{l}{\emph{Injection axis}}\\
\;Uniform: $+$ASR          & 200 &  9 &  6 &  7.5\% & 5.4 \\
\;Cascade: $+$ASR          & 200 & 10 &  4 &  7.0\% & 5.2 \\
\;ASR$+$OCR vs ASR         & 200 &  1 &  4 &  2.5\% & 3.1 \\
\midrule
\multicolumn{6}{l}{\emph{Combined}}\\
\;Cascade$+$ASR vs unif.   & 200 &  28 &  15 & 21.5\% & 9.2 \\
\;Cascade$+$ASR vs unif.   & 900 & 114 &  99 & 23.7\% & 4.5 \\
\bottomrule
\end{tabular}
\end{table}

\subsection*{Is the subset--full-bucket gap sampling noise?}
\label{app:substab}
The main paper reports that our $n{=}200$ estimates are $2$--$4\times$
larger than the $n{=}900$ estimates of the same contrast. Two
explanations compete: the subset is systematically unrepresentative, or a
$200$-item paired estimate is simply too noisy to be reported as an
effect. We can separate them without running any inference, because the
$n{=}200$ item set is a deterministic \emph{prefix} of the $n{=}900$ set
(the loader truncates the benchmark in its native order, so the subset
covers $67$ of the $300$ long videos), and the full-bucket predictions
therefore already contain the subset.

We draw $B{=}20{,}000$ random $200$-item subsets without replacement from
the $900$ paired outcomes and recompute the delta and the exact McNemar
$p$ on each (\texttt{scripts/subset\_stability.py}, seed
$20270101$); Table~\ref{tab:substab-full} summarises the result. The
resampling distribution contains the paper's own prefix estimate
comfortably---$+3.5$\,pt sits at the $82$nd percentile and $+6.5$\,pt at
the $94$th---so neither is an outlier, and the gap is sampling
variability rather than a pathological subset.

The width of that distribution is the actual finding. A $200$-item draw
reverses the sign of the full-bucket effect $32$--$36\%$ of the time, and
its $95\%$ range spans about $10.5$\,pt, wider than the entire range of
effect sizes this literature reports. At the same time only $2.8\%$ of
draws reach nominal $p{<}0.05$, consistent with an MDE of $8.5$\,pt and
a full-bucket effect of $1.2$\,pt. In this regime a subset-scale
comparison is close to a coin flip on sign and nearly silent on
significance.

\begin{table}[h]
\centering
\caption{Repeated subsampling of $200$ items from the $n{=}900$ paired
outcomes ($B{=}20{,}000$ draws; post-processing of cached predictions
only, no new inference).}
\label{tab:substab-full}
\small
\setlength{\tabcolsep}{4pt}
\begin{tabular}{lcc}
\toprule
 & Cascade & Cascade$+$ASR \\
 & vs unif. & vs unif. \\
\midrule
Full-bucket $\Delta$ ($n{=}900$)    & $+1.2$ & $+1.7$ \\
Paper's prefix $\Delta$ ($n{=}200$) & $+3.5$ & $+6.5$ \\
\midrule
Median over $200$-subsets           & $+1.5$ & $+1.5$ \\
$95\%$ range                        & $[-4.0,+6.5]$ & $[-4.5,+7.5]$ \\
Standard deviation                  & $2.8$ & $3.0$ \\
Sign flips vs full bucket           & $35.8\%$ & $31.9\%$ \\
$\Pr(\Delta \ge +5$\,pt$)$          & $8.3\%$ & $13.0\%$ \\
Draws reaching $p{<}0.05$           & $2.8\%$ & $3.5\%$ \\
Percentile of paper's prefix        & $81.8$ & $93.6$ \\
\bottomrule
\end{tabular}
\end{table}

\subsection*{Cross-model context}
The pipeline is VLM-agnostic by construction; headline numbers use
Qwen2.5-VL-7B and the selection-only null is checked on a same-code-path
Qwen2-VL-7B backbone (\emph{Cross-Backbone Robustness} section above). For a controlled selector
comparison, the AKS port consumes identical SigLIP scores and
differs only in the selection rule; under this control it scores
$48.1\%$ vs.\ uniform $46.2\%$ ($p{=}0.27$, n.s.), consistent with the F1
prediction that a selection-rule change alone leaves accuracy flat in
this regime. For published-number context, VideoTree~\cite{wang2024videotree}
reports $54.2$ on Video-MME long but uses a GPT-4 reasoner with dense
captions; AKS~\cite{tang2025aks} reports $+0.9$ to $+2.3$\,pt
selection-only gains on Video-MME overall. Both live on the selection
axis, which F1 finds does not separate at full scale in our matched
setting.
Table~\ref{tab:protocol} summarises why absolute cross-paper numbers
are not comparable and which variables each method controls.

\begin{table}[h]
\centering
\caption{\textbf{Why absolute cross-paper numbers are not comparable.}
Published systems differ on training, reasoner capacity, frame budget,
and scoring protocol, so a leaderboard number cannot isolate the
selection/injection/expert routing decision \name{} studies. We
therefore run \emph{controlled} matched-backbone, matched-budget
contrasts (AKS reproduced under our exact SigLIP features and $K{=}8$),
rather than competing on an entangled leaderboard.}
\label{tab:protocol}
\small
\resizebox{\columnwidth}{!}{%
\begin{tabular}{lcccc}
\toprule
Method & Train-free & Reasoner & Frames & Scoring \\
\midrule
TSPM~\cite{li2024tspm} (AVQA)       & $\times$ & specialist  & 60      & closed/exact \\
VideoTree~\cite{wang2024videotree}  & \checkmark & GPT-4       & dense   & open \\
AKS~\cite{tang2025aks}              & \checkmark & 7B          & 8       & open \\
\textbf{\name{} (ours)}          & \checkmark & 7B          & 8       & open$+$judge \\
\bottomrule
\end{tabular}}
\end{table}

\textbf{Cross-benchmark (EgoSchema).} On EgoSchema ($n{=}500$, $K{=}8$,
Qwen2.5-VL-7B) the audit-profile cascade scores $66.2\%$ against $67.8\%$
for uniform sampling ($-1.6$\,pt, $95\%$ CI $[-4.6,+1.4]$, $b/c{=}24/32$,
$p{=}0.35$; $90\%$ CI $[-4.0,+0.8]$, equivalent within $\pm5$ but not
$\pm3$\,pt); the early-pipeline run gave the same $66.2$ vs.\ $67.8\%$
($b/c{=}27/35$, $p{=}0.37$), although the two cascade runs disagree on
$60$ of the $500$ questions ($30$ in each direction), so the equal
accuracies are partly coincidental.
Neither run shows a gain from selection. The three ported selectors, run
through the same harness (Table~\ref{tab:egoports}), do not help either:
AKS $-2.6$\,pt, KeyScore $-4.0$\,pt and VideoTree $-0.2$\,pt against
uniform. Only VideoTree and the cascade are equivalent to uniform within
$\pm5$\,pt; the KeyScore port is nominally \emph{below} it ($95\%$ CI
$[-7.8,-0.2]$, $p{=}0.049$), which does not survive a correction for the
four selector contrasts, and the AKS interval is wider than the margin.

\begin{table}[h]
\centering
\caption{\textbf{EgoSchema, $n{=}500$, $K{=}8$, Qwen2.5-VL-7B.} $\Delta$ is
against uniform with its $95\%$ paired-bootstrap interval; $p$ is exact
McNemar, unadjusted. $^\ddagger$Equivalent to uniform within $\pm5$\,pt
(TOST on the $90\%$ CI).}
\label{tab:egoports}
\small
\setlength{\tabcolsep}{4pt}
\begin{tabular}{lcccc}
\toprule
Method & Acc.\ (\%) & $\Delta$ & $95\%$ CI & $p$ \\
\midrule
Uniform-$K$                   & 67.8 & ---    & ---             & --- \\
Cascade$^\ddagger$            & 66.2 & $-1.6$ & $[-4.6,+1.4]$   & 0.350 \\
AKS (port)                    & 65.2 & $-2.6$ & $[-6.2,+0.8]$   & 0.182 \\
KeyScore (port)               & 63.8 & $-4.0$ & $[-7.8,-0.2]$   & 0.049 \\
VideoTree (port)$^\ddagger$   & 67.6 & $-0.2$ & $[-3.6,+3.2]$   & 1.000 \\
\bottomrule
\end{tabular}
\end{table}

\subsection*{Extended F3(b): expert-routing analysis}

\begin{table}[h]
\centering
\caption{\textbf{Per-type MUSIC-AVQA breakdown} on the full split
($n{=}9{,}129$, both models at $K{=}8$, same questions aligned by
\texttt{qid}, same LLM judge). \emph{Audio-blind} is the visual-only
Qwen2-VL-7B; \emph{AV expert} routes the same question to
VideoLLaMA2-AV~\cite{cheng2024videollama2}. $p$ is two-sided exact
McNemar. Every type improves, including the purely visual ones, so the
swap is not an audio ablation (the silenced-soundtrack control is below).}
\label{tab:avexpert_pertype}
\small
\resizebox{\columnwidth}{!}{%
\begin{tabular}{lccccc}
\toprule
Question type & $n$ & Audio-blind & \textbf{AV expert} & $\Delta$ & $p$ \\
\midrule
Visual-Counting        & 1197 & 83.0 & \textbf{89.1} & $+6.1$  & $1.9\times10^{-8}$ \\
Visual-Location        & 1225 & 65.7 & \textbf{93.1} & $+27.3$ & $7.3\times10^{-75}$ \\
Audio-Counting         & 1017 & 85.6 & \textbf{88.1} & $+2.5$  & $0.013$ \\
Audio-Comparative      & 594  & 52.2 & \textbf{69.0} & $+16.8$ & $2.6\times10^{-10}$ \\
AV-Existential         & 988  & 81.1 & \textbf{83.8} & $+2.7$  & $0.014$ \\
AV-Counting            & 1265 & 66.7 & \textbf{81.0} & $+14.3$ & $1.8\times10^{-20}$ \\
AV-Comparative         & 1101 & 52.6 & \textbf{69.2} & $+16.6$ & $1.3\times10^{-16}$ \\
AV-Location            & 920  & 57.1 & \textbf{76.2} & $+19.1$ & $6.1\times10^{-22}$ \\
AV-Temporal            & 822  & \textbf{37.6} & \textbf{70.9} & $+33.3$ & $2.8\times10^{-54}$ \\
\midrule
Micro-average (all)    & 9129 & 66.1 & \textbf{81.2} & $+15.1$ & $1.5\times10^{-167}$ \\
Macro-average (types)  & ---  & 64.6 & \textbf{80.1} & $+15.4$ & --- \\
\bottomrule
\end{tabular}}
\end{table}

\textbf{Silenced-soundtrack control.} To separate audio from the model
swap we ran VideoLLaMA2.1-7B-AV a second time with its waveform replaced
by digital silence (a zero waveform, the model's own no-audio input) on
$200$ questions of each type, drawn once with a fixed seed ($1{,}800$
requested, $1{,}799$ usable: one video has no audio track, so there is
nothing to silence). The audio-blind and audio-on arms are the full-split
runs restricted to the same questions, and one judge grades all three arms
(Table~\ref{tab:silent}).
Because the judge sees only the question, the gold answer and the
predicted answer and decodes greedily, $1{,}631$ of the $1{,}799$
silenced answers reused the verdict of an identical, already judged
audio-on or audio-blind answer; the other $168$ were judged anew, and
re-judging $30$ reused verdicts reproduced all $30$. Silencing costs
$1.3$\,pt (paired bootstrap $95\%$ CI $[-0.1,+2.6]$) of the $+15.1$\,pt
swap (micro-averages post-stratified to the full split's type mix; the
swap on the subset equals the full split's), so $8\%$ of the gain
(CI $[-0.5,16.4]$) is audio and $92\%$ is not. The in-loop string
matcher gives the same picture ($+16.8$\,pt swap, $+1.4$ $[+0.1,+2.7]$
audio, $8\%$ $[0.5,15.7]$). Per type, no audio effect is individually
significant (exact McNemar $p{\ge}0.085$ at $n{=}200$; the largest are
Audio-Comparative, $+6.5$ $[-0.5,+13.0]$, and AV-Comparative, $+5.0$
$[-1.0,+11.5]$), and AV-Temporal gains $+25.1$ of its $+27.6$\,pt
without sound.

\textbf{Mismatched-soundtrack control.} Silence is a neutral input; to test
whether the expert reads the soundtrack's \emph{content} we also gave each
video the soundtrack of a different video (a fixed shuffle of the $6{,}399$
videos with no video keeping its own, so every video lends and receives one;
same $1{,}800$ questions, no failures; the swapped tensors differ from the
originals and have the same shape). Post-stratified and judge-graded, the
arms score $67.0$ (audio-blind), $77.5$ (mismatched), $80.9$ (silenced) and
$82.1\%$ (true soundtrack). The wrong soundtrack is $3.4$\,pt below
silence ($[-4.8,-2.0]$, $b/c{=}66/128$, $p{=}10^{-5}$) and $4.7$\,pt below the
true soundtrack ($[+3.2,+6.2]$, $p{=}10^{-9}$), so the expert does use the
audio, and can be misled by it; the right audio adds $1.3$\,pt over
silence ($[-0.1,+2.6]$). Against the audio-blind model even the wrong
soundtrack leaves $+10.5$\,pt ($[+8.1,+12.9]$), $69\%$ of the swap, versus
$92\%$ with silence; the in-loop matcher gives the same numbers within
$1.6$\,pt. The share of the swap that is audio therefore depends on the
counterfactual (silence: $8\%$; wrong audio: $31\%$), and it is at most
about a third. This is one expert, one random assignment of soundtracks and
a $\pm3$\,pt resolution per type, and the expert's in-domain MUSIC-AVQA
training makes it likely that the visual stream suffices for most questions.

\begin{table}[h]
\centering
\caption{\textbf{Same expert, soundtrack silenced} (VideoLLaMA2.1-7B-AV,
$K{=}8$; $200$ questions per type, same questions in all three arms, same
judge). \emph{AV$-$blind} is the swap and \emph{AV$-$silenced} the part of
it that is audio. The micro row weights types by their share of the full
split.}
\label{tab:silent}
\small
\setlength{\tabcolsep}{3pt}
\resizebox{\columnwidth}{!}{%
\begin{tabular}{lcccccc}
\toprule
Question type & $n$ & Blind & Silenced & AV & AV$-$blind & AV$-$silenced \\
\midrule
Visual-Counting & 200 & 83.5 & 90.0 & 89.5 & $+6.0$ & $-0.5$ \\
Visual-Location & 200 & 61.0 & 95.0 & 95.0 & $+34.0$ & $+0.0$ \\
Audio-Counting & 200 & 83.5 & 85.5 & 86.0 & $+2.5$ & $+0.5$ \\
Audio-Comparative & 200 & 56.0 & 62.5 & 69.0 & $+13.0$ & $+6.5$ \\
AV-Existential & 200 & 83.0 & 85.5 & 87.0 & $+4.0$ & $+1.5$ \\
AV-Counting & 200 & 67.0 & 83.0 & 83.0 & $+16.0$ & $+0.0$ \\
AV-Comparative & 200 & 59.5 & 67.5 & 72.5 & $+13.0$ & $+5.0$ \\
AV-Location & 200 & 56.5 & 77.5 & 76.0 & $+19.5$ & $-1.5$ \\
AV-Temporal & 199 & 41.7 & 66.8 & 69.3 & $+27.6$ & $+2.5$ \\
\midrule
Micro (post-stratified) & 1799 & 67.0 & 80.9 & 82.1 & $+15.1$ & $+1.3$ \\
\bottomrule
\end{tabular}}
\end{table}

\textbf{A deployable text-only dispatcher.} Table~\ref{tab:avexpert_pertype}
reports the \emph{headroom} of expert routing under a gold-type oracle.
We now ask the deployable question: \emph{if a single-pass dispatcher had
to choose the expert from the question text alone---no gold question-type
label---how close to that headroom does it get?} We apply the planner's
fixed keyword rule (audio/speech cue words, mirroring
\texttt{cascade/planner.py}) to each question, route to the audio expert
iff a cue fires, and read the answer of the selected expert from the two
pre-computed passes aligned by \texttt{qid}. This is exactly the
input--output behaviour of a live single-pass dispatcher; the
pre-computation only saves wall-clock, not information. On the full
$n{=}9{,}129$ split the text-only dispatcher reaches \textbf{$78.6\%$}---%
$2.6$\,pt below the always-on audio expert ($81.2\%$) and $+12.5$\,pt over the
audio-blind base ($66.1\%$; exact McNemar $p{=}6.6\!\times\!10^{-130}$,
$95\%$ bootstrap CI $[+11.5,+13.5]$), capturing $57\%$ of the oracle
headroom \emph{without} access to any gold label. Expert routing is thus
deployable for \emph{accuracy}, not merely an oracle bound.

\textbf{Cost is not yet recovered (honest scope).} The same trace exposes
the limit. The keyword rule routes $84.9\%$ of questions to the expensive
audio expert (precision $0.22$, recall $0.86$ against the
``audio-actually-helped'' set), because MUSIC-AVQA is overwhelmingly
audio-visual---almost every question mentions an audio cue. So on this
benchmark the dispatcher is accuracy-preserving but \emph{not}
cost-saving: it behaves close to always-on. A genuinely cost-aware
dispatcher would need to predict the per-item event ``the base model will
fail \emph{and} audio will fix it,'' which is entangled with base-model
competence and is not a property of the question text (the
$1{,}990$ helped items are diffuse across every question type, each of
which also contains items the audio expert \emph{hurts}). Recovering cost
therefore requires a mixed-modality benchmark with a non-trivial fraction
of audio-irrelevant questions to gate on; we leave cost-aware dispatch on
such a benchmark to future work and make no cost-saving claim here.

\textbf{Significance.} Every per-type contrast in
Table~\ref{tab:avexpert_pertype} is significant at the $0.05$ level and
seven of nine at $p{<}10^{-7}$. The headline cells: AV-Temporal gains
$+33.3$\,pt (exact McNemar $p{=}2.8\!\times\!10^{-54}$; bootstrap $95\%$
CI $[+29.6,+37.1]$\,pt), the micro-average $+15.1$\,pt
($p{=}1.5\!\times\!10^{-167}$, CI $[+14.0,+16.1]$) and the pooled
audio-visual types $+16.5$\,pt ($p{=}1.6\!\times\!10^{-96}$, CI
$[+15.0,+18.1]$). We stress that the purely visual types also move
($+6.1$ and $+27.3$\,pt), which is why the main paper reads this axis as
a model swap that upper-bounds, rather than isolates, the value of audio
evidence.

\textbf{SOTA context (directional, not controlled).} For the hardest
AV-Temporal type, specialist audio-visual systems---trained end-to-end
with a closed $42$-way answer vocabulary and exact-match scoring---report
$65.8$ (AVST~\cite{li2022musicavqa}), $69.0$ (PSTP-Net~\cite{li2023pstpnet}),
$68.0$ (LAVISH~\cite{lin2023lavish}), and $71.2$ (TSPM~\cite{li2024tspm},
the non-LLM SOTA). Our routed open-ended generation lands at $70.9$ on
the same type. This is directional context, not an apples-to-apples
comparison: those systems classify over a fixed vocabulary scored by
exact match, whereas \name{} generates free-form answers scored by an
LLM judge. The directional comparison shows that answering-model choice
can approach reported specialist accuracy, but the protocols are not
commensurate.

\textbf{Routing from question text alone.} A deployable router that
decides the expert from the \emph{question text only} (planner's
audio/speech keyword cue, no gold type label) sends $84.9\%$ of
questions to the audio expert and reaches $78.6\%$ micro---$2.6$\,pt
short of the always-audio number ($81.2\%$) and $+12.5$\,pt over the
audio-blind backbone ($p{=}6.6\!\times\!10^{-130}$). MUSIC-AVQA is
modality-homogeneous, so a per-question router largely collapses to ``use the
audio expert for this dataset''; the per-question \emph{cost-saving}
value of routing can only be shown on a modality-mixed benchmark. On
MUSIC-AVQA the contribution is the routing \emph{headroom} (a per-item
oracle reaches $87.9\%$, leaving $+6.7$\,pt above always-audio).

\begin{figure}[h]
\centering
\includegraphics[width=\columnwidth]{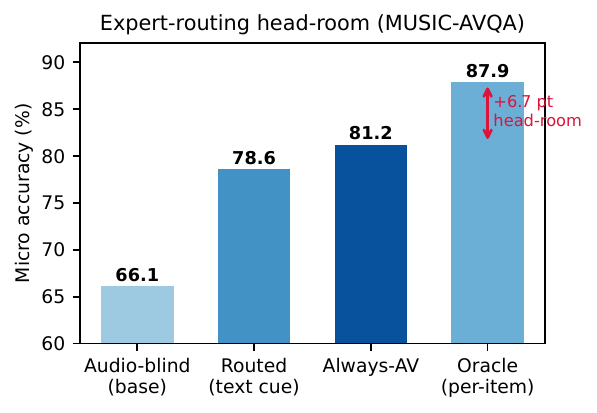}
\caption{\textbf{Expert-routing headroom (F3(b)).} A deployable router
reading the \emph{question text only} (Routed, $78.6$) recovers most of
the always-audio accuracy ($81.2$) and $+12.5$\,pt over the
audio-blind base ($66.1$). A per-item oracle reaches $87.9$, leaving
$+6.7$\,pt of headroom above always-audio.}
\label{fig:routingladder}
\end{figure}

\clearpage
{
    \small
    \bibliographystyle{ieeenat_fullname}
    \bibliography{refs}
}

\end{document}